\PassOptionsToPackage{dvipsnames}{xcolor}
\documentclass[journal]{IEEEtran}
\IEEEoverridecommandlockouts

\usepackage{amsmath}
\allowdisplaybreaks
\usepackage{amssymb}
\usepackage{amsthm}
\usepackage{bm}
\usepackage{mathtools}
\usepackage{times}
\usepackage{bbm}

\usepackage{graphicx}
\usepackage{setup/support-caption}
\usepackage[font=footnotesize]{subcaption}
\usepackage{floatrow}
\usepackage{xspace}
\usepackage{blindtext}
\usepackage{algorithm} 
\usepackage{algorithmic}
\usepackage{tabularx}
\usepackage{booktabs}
\usepackage{orcidlink}
\usepackage{xcolor}
\usepackage{makecell}

\usepackage{hyperref}
\hypersetup{
colorlinks=true,
linkcolor=blue,
filecolor=blue,      
urlcolor=blue,
citecolor=blue
}
\usepackage{tcolorbox}
\usepackage{etoolbox}

\definecolor{michiganblue}{rgb}{0,0.153,0.310}
\definecolor{michiganmaize}{rgb}{1.0,0.796,0.020}
\definecolor{darkblue}{rgb}{0,0,0.7}

\DeclareMathOperator*{\argmin}{arg\,min}
\DeclareMathOperator*{\argmax}{arg\,max}

\long\def\comment#1{}

\newcommand{\xmath}[1] {\ensuremath{#1}\xspace}

\newcommand{\blmath}[1] {\xmath{\bm{#1}}}
\newcommand{\paren}[1] {\xmath{\left(#1\right)}}

\newcommand{\bmat}[1] {\xmath{\left[\begin{matrix} #1 \end{matrix}\right]}}
\newcommand{\normfrob}[1] {\xmath{\left\| #1 \right\|_{\mathrm{F}}}} 
\newcommand{\normfrobr}[1] {\xmath{\| #1 \|_{\mathrm{F}}}} 

\newcommand{\pbox}[1] {%
\makebox[0pt][r]{\raisebox{7mm}[0pt][0pt]{\small #1}}\ignorespaces}
\newcommand{\be} {\begin{equation}}
\newcommand{\ee}[1] {\label{#1}\end{equation}\pbox{#1}}
\newcommand{\eref}[1] {(\ref{#1})}

\newcommand\highlightReference[1]{%
  \expandafter\newcommand\csname highlightReference-#1\endcsname{}%
}
\let\oldbibitem\bibitem
\def\bibitem#1 #2\par{%
  \expandafter\ifx\csname highlightReference-#1\endcsname\relax
    \oldbibitem{#1}#2\par
  \else
    \oldbibitem{#1}\highlight{#2}\par
  \fi
}

\newcommand\highlight[1]{\textcolor{blue}{#1}}

\newcommand{\supnew} {^{\mathrm{new}}}

\newcommand{\green}[1]{\color{OliveGreen}{#1} \color{black}}
\newcommand{\fref}[1] {Fig.~\ref{#1}}
\newcommand{\aref}[1] {Alg.~\ref{#1}}

\newcommand{\sref}[1] {Sec.~\ref{#1}}
\newcommand{\tref}[1] {Thm.~\ref{#1}}
\newcommand{\tabref}[1] {Table~\ref{#1}}
\newcommand{\defequ}{\triangleq}

\newcommand{\bs}{\begin{equation}\begin{split}}

\newcommand{\st}{\hspace{2mm} \text{s.t.} \hspace{2mm}}

\newcommand{\A}{\blmath{A}}

\newcommand{\D}{\blmath{D}}

\newcommand{\I}{\blmath{I}}
\renewcommand{\L}{\blmath{L}}
\newcommand{\R}{\blmath{R}}
\newcommand{\U}{\blmath{U}}
\newcommand{\V}{\blmath{V}}
\newcommand{\Q}{\blmath{Q}}
\newcommand{\W}{\blmath{W}}
\newcommand{\Y}{\blmath{Y}}
\newcommand{\M}{\blmath{M}}
\newcommand{\Z}{\blmath{Z}}

\newcommand{\Sig}{\blmath{\Sigma}}

\newcommand{\x}{\blmath{x}}

\newcommand{\bnu}{\blmath{\nu}}

\renewcommand{\H}{\blmath{H}}

\newcommand{\X}{\blmath{X}}
\newcommand{\y}{\blmath{y}}
\newcommand{\z}{\blmath{z}}
\newcommand{\e}{\blmath{e}}
\renewcommand{\r}{\blmath{r}}

\newcommand{\Pb}{\blmath{P}}

\newcommand{\cA}{\xmath{\mathcal{A}}}
\newcommand{\cB}{\xmath{\mathcal{B}}}
\newcommand{\cC}{\xmath{\mathcal{C}}}
\newcommand{\cL}{\xmath{\mathcal{L}}}
\newcommand{\cR}{\xmath{\mathcal{R}}}
\newcommand{\Ck}{\xmath{\cC_k}}

\newcommand{\real}{\mathbb{R}}

\renewcommand{\xi}{\xmath{x_i}} 
\newcommand{\yi}{\xmath{\y_i}}

\let\originalPi=\Pi 
\renewcommand{\Pi}{\blmath{\originalPi}}

\newcommand{\Sb}{\blmath{S}}

\newcommand{\0}{\blmath{0}} 
\newcommand{\bepsilon}{\boldsymbol{\epsilon}}

\newtheorem{theorem}{Theorem}

\usepackage{microtype}

\begin{document}

\title{
ALPCAHUS: Subspace Clustering
\\
for Heteroscedastic Data
}

\author{
\IEEEauthorblockN{ 
Javier Salazar Cavazos\orcidlink{0009-0009-1218-9836},~\IEEEmembership{Graduate Student Member,~IEEE}, \\
Jeffrey A. Fessler\orcidlink{0000-0001-9998-3315},%
~\IEEEmembership{Fellow,~IEEE}, and 
Laura Balzano\orcidlink{0000-0003-2914-123X},%
~\IEEEmembership{Senior Member,~IEEE}
\\}
\IEEEauthorblockA{
Electrical \& Computer Engineering (ECE)} Department, University of Michigan, Ann Arbor, Michigan, United States\\
Email: \texttt{\{javiersc, fessler, girasole\}@umich.edu }}

\maketitle

\begin{abstract}
Principal component analysis (PCA) is a key tool
in the field of data dimensionality reduction.
Various methods have been proposed
to extend PCA to the union of subspace (UoS) setting
for clustering data that comes from multiple subspaces
like $K$-Subspaces (KSS).
However, some applications
involve heterogeneous data that vary in quality
due to noise characteristics associated with each data sample.
Heteroscedastic methods aim to deal with such mixed data quality.
This paper develops a heteroscedastic-based subspace clustering method,
named ALPCAHUS,
that can estimate the sample-wise noise variances
and use this information
to improve the estimate of the subspace bases
associated with the low-rank structure of the data.
This clustering algorithm builds on K-Subspaces (KSS) principles
by extending the recently proposed heteroscedastic PCA method, named LR-ALPCAH,
for clusters with heteroscedastic noise in the UoS setting.
Simulations and real-data experiments show the effectiveness
of accounting for data heteroscedasticity
compared to existing clustering algorithms.
Code available at
\url{https://github.com/javiersc1/ALPCAHUS}.
\end{abstract}

\begin{IEEEkeywords}
Heteroscedastic data, heterogeneous data quality, subspace bases estimation,  
subspace clustering, union of subspace model, unsupervised learning.
\end{IEEEkeywords}

\section{Introduction}

Many modern data science problems require
learning an approximate signal subspace basis
for some collection of data.
This is important for downstream tasks
involving subspace basis coefficients such as
classification \cite{classification},
regression \cite{regression},
and compression \cite{compression}.
Besides subspace learning, one may be interested in clustering data points
that originate from multiple subspaces. 
Formally, subspace clustering, or union of subspace (UoS) modeling, 
is an unsupervised machine learning problem
where the goal is to cluster unlabeled data
and find the subspaces associated with each data cluster.
When the cluster assignments are known,
it is easy to find the subspaces,
and vice versa.
This problem becomes nontrivial
when both components must be estimated
\cite{subspace_review}. 
This clustering problem has many applications,
such as
image segmentation \cite{image_segmentation},
motion segmentation \cite{motion_segmentation}, 
image compression \cite{image_compression}, and 
system identification \cite{systems_identification}.

\begin{figure}
  \centering
  \includegraphics[width=0.97\textwidth]{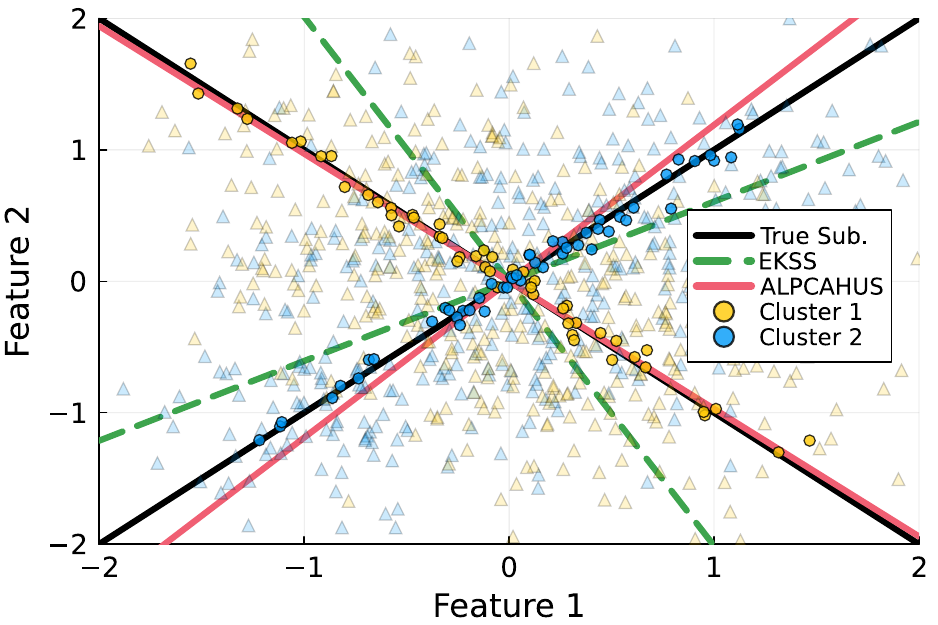}
  \caption{Two 1D subspaces, colored blue and yellow,
  with data consisting of two noise groups shown
  with circle (low noise) and triangle (high noise) markers.}
  \label{fig:uos_example}
\end{figure}

Some applications
involve heterogeneous data samples that vary in quality
due in part to noise characteristics associated with each sample. A few examples of heteroscedastic datasets
include environmental air quality data \cite{epa},
astronomical spectral data \cite{astronomical_data},
and biological sequencing data \cite{bacteria_data}.
In heteroscedastic settings,
the noisier data samples can significantly corrupt the basis estimates \cite{asymptotic_pca}.
In turn, this corruption can worsen clustering performance as seen in \fref{fig:uos_example}.
Popular clustering methods such as
Sparse Subspace Clustering (SSC) \cite{ssc}, 
$K$-Subspaces (KSS) \cite{k-subspaces},
and Subspace Clustering via Thresholding (TSC) \cite{heckel2015robust}, all implicitly assume that data quality is consistent. For example, in SSC, the method relies on the self-expressiveness property of data
that uses other similar samples to estimate each single sample.
From our experiments, we found that this implicit data quality assumption
can degrade clustering quality
for heteroscedastic data.

Because of these limitations, 
we developed a subspace clustering algorithm, 
inspired from KSS principles, that
explicitly models noise variance terms, 
without assuming that data quality is known. 
The method adaptively clusters data while learning noise characteristics. 
See \fref{fig:uos_example} for a visualization
where Ensemble KSS (EKSS) \cite{ekss} returns poor subspace bases estimates
whereas our method found more accurate subspace bases
and improved clustering quality.

We extend our previous work \cite{alpcah_journal}
by generalizing the LR-ALPCAH formulation to the UoS setting
for clustering heteroscedastic data.
The proposed approach
achieved ${\sim}3$ times lower clustering error than existing methods,
and it achieved a relatively low clustering error
even when very few high quality samples were available.
The paper is divided into a few key sections.
\sref{paper:problem}
introduces the heteroscedastic problem formulation
for subspace clustering.
\sref{paper:works} discusses related work in subspace clustering
and reviews the heteroscedastic subspace algorithm LR-ALPCAH.
\sref{paper:method}
introduces the proposed subspace clustering method named ALPCAHUS. 
\sref{paper:experiments}
covers synthetic and real data experiments
that illustrate the effectiveness of modeling heteroscedasticity in a clustering context.
Finally, \sref{paper:conclusion}
discusses some limitations of our method and possible extensions.
\section{Problem Formulation}

\label{paper:problem}
Let $K$ denote the number of subspaces that is either known beforehand
or estimated using other methods.
Before describing the general union of subspace model, the single subspace model under $K = 1$ is introduced.

\subsection{Single Subspace Model \texorpdfstring{($K=1$)}{(K=1)}}

Let $\yi \in \mathbb{R}^{D}$ denote the data samples
for index $i \in \{1,\ldots,N \}$,
where $D$ denotes the ambient dimension and $N$ is the total number of samples.
Let $\x_i$ represent the low-dimensional data sample
generated by $\x_i = \U \z_i$
where $\U \in \mathbb{R}^{D \times d}$ is an unknown basis
for a subspace of dimension $d$
and $\z_i \in \mathbb{R}^d$ are the corresponding basis coordinates.
Then the heteroscedastic model we consider is
\begin{equation}
\yi = \x_i + \bepsilon_i
\quad \text{where} \quad
\bepsilon_i \sim \mathcal{N}(\0, \nu_i \I)
\label{eq:heteroscedastic_subspace_model}
\end{equation}
assuming Gaussian noise with variance $\nu_i$,
where \I denotes the $D \times D$ identity matrix.
In this work, we consider the case where each data sample may have its own noise variance, however, one can adapt this method to consider the case where there are
$G$ groups of data having shared noise variance terms
$\{ \nu_1,\ldots,\nu_G \}$.
\sref{sec:works_pca}
discusses an optimization problem based on this model that estimates the heterogeneous noise variances
$\{\nu_i\}_{i=1}^N$
and the subspace basis \U. 

\subsection{Union of Subspaces Model \texorpdfstring{($K \geq 1$)}{(K > 1)}}

Let
$\Y = \bmat{\y_1 & \ldots & \y_N} \in \mathbb{R}^{D \times N}$
denote a matrix whose columns consist of all $N$ data points
$\y_i \in \mathbb{R}^D$.
We generalize
\eref{eq:heteroscedastic_subspace_model}
to model the data
with a union of subspaces model by
\begin{align}
\y_i & = \x_i  + \bepsilon_i
\nonumber \\
\x_i & = \U_{k_i} \z_i
\text{ for some }
k_i \in \{1, \ldots, K \}
\label{eq:uos_heteroscedastic_model},
\end{align}
where
$\U_k \in \mathbb{R}^{D \times d_k}$ is a subspace basis that
has subspace dimension $d_k$.
Here, $\z_i \in \mathbb{R}^{d_k}$
denotes the basis coefficients associated with $\x_i$,
and $\bepsilon_i \in \mathbb{R}^D$ denotes noise for that point
drawn from $\bepsilon_i \sim \mathcal{N}(\0, \nu_i \I)$.

If the subspace bases were known,
then one would like to find the associated subspace label
$c_i \in \{ 1, \ldots,K \}$ for each data sample
by solving the following optimization problem
\begin{equation}
    c_i = \argmin_{k} \| \y_i -  \U_k \U_k' \y_i \|_2^2,
    \quad \forall \y_i \in \Y.
\end{equation}
This label describes the subspace association of a data point $\y_i$
that has the lowest residual between the original sample and its reconstructed value
given the subspace.
In general,
the goal is to estimate all of the subspace bases
$\mathcal{U} = ( \U_1, \ldots, \U_K )$
and cluster assignments
$\cC = (c_1, \ldots, c_N)$ associated with all the data samples simultaneously.

\section{Related Works}
\label{paper:works}

\subsection{Subspace Clustering}
\label{paper:works_clustering}

Many subspace clustering algorithms fall into
a general umbrella of categories such as
algebraic methods \cite{algebraic_method},
iterative methods \cite{iterative_method}, 
statistical methods \cite{statistical_method},
and spectral clustering methods \cite{spectral_method1} \cite{spectral_method2}.
In recent years, both spectral clustering-based methods and iterative methods
have become popular.

\subsubsection{Spectral Clustering}
\label{paper:works_spectral_clustering}
Many methods build on spectral clustering.
This method is often used for clustering nodes in graphs.
Spectral clustering aims to find the minimum cost ``cuts'' in the graph
to partition nodes into clusters. Given some collection of data points, one way to construct a graph is to assume that ``nearby'' data samples are highly connected in the graph. Thus, it is possible to construct a graph from data samples, with various levels of connectedness, by using some metric to assign affinity/similarity for each pair of points. 
Let $\mathcal{G}(\mathcal{V}, \W)$ correspond to a graph
that consists of vertices $\mathcal{V} = \{v_1, \ldots, v_N\}$
and edge weights $\W \in \mathbb{R}^{N \times N}$
such that $w_{ij}$ corresponds to some nonnegative weight,
or similarity, between $v_i$ and $v_j$.
The graph $\mathcal{G}$ is assumed to be undirected,
i.e., $\W=\W'$.
The degree of a node is defined as
$d_i = \sum_{j=1}^N w_{ij}$ 
and can be collected to form a degree matrix
such that $\D = \text{diag}(d_1,\ldots,d_N)$.
Let $\cA_1,\ldots,\cA_K$ form a $K$-partition on $\mathcal{G}(\mathcal{V},\W)$
and $\text{cut}(\cA,\cB) = \sum_{i \in \cA, j \in \cB} w_{ij}$.
Then, spectral clustering aims to solve the following
\begin{equation}
    \argmin_{\cA_1,\ldots,\cA_K}
    \frac{1}{2} \sum_{i=1}^K \frac{\text{cut}(\cA_i, \bar{\cA}_i)}{|\cA_i|}
    \label{eq:ratio_cut}
\end{equation}
where $\bar{\cA}_i$ is all vertices not belonging in $\cA_i$
and $|\cA_i|$ is the number of vertices belonging to the $i$th partition.
Instead of working with $\W$ directly,
one computes a normalized graph Laplacian such as
$ \L = \I - \D^{-1} \W $ to make the influence of heavy degree nodes more similar to low degree nodes.
Let $h_{i,j}$ indicate the $j$th cluster membership for the $i$th point by
\begin{equation}
h_{ij} = \left\{
\begin{array}{ll}
1/\sqrt{|\cA_j|} & \text{ if } v_i \in \cA_j
\\
0 & \text{ otherwise.}
\end{array}
\right.
\label{eq:ratio_cut2}
\end{equation}
Collecting the indicator values into a matrix
$\H \in \mathbb{R}^{N \times K}$,
one can rewrite
\eref{eq:ratio_cut} in matrix notation as 
\begin{equation}
    \frac{1}{2} \sum_{i=1}^K \frac{\text{cut}(\cA_i, \bar{\cA}_i)}{|\cA_i|}
    = \sum_{i=1}^K (\H' \L \H)_{ii}
    = \text{Tr}(\H' \L \H),
    \label{eq:ratio_cut3}
\end{equation}
where $\text{Tr}(\H) = \sum_i \H_{ii}$ denotes the trace of a matrix.
The indicator vectors are discrete,
which makes the problem computationally challenging.
One way to relax the problem is to allow arbitrary real values for $\H$
and instead solve
\begin{equation}
    \hat{\H} = \argmin_{\H \in \real^{N \times K}}
    \text{Tr}(\H' \L \H) \text{ s.t. } \H' \H = \I
    \label{eq:spectral_clustering}
\end{equation}
where the notation s.t. means subject to some conditions. Trace minimization problems with semi-unitary constraints
are well-studied in the literature,
and it is easy to see that $\hat{\H}$
consists of the first $K$ eigenvectors of $\L$ associated with the smallest eigenvalues, i.e., 
$\lambda_1(\L) \leq \ldots \leq \lambda_N (\L)$
where $\lambda_i(\L)$ denotes the $i$th eigenvalue of $\L$.
Once $\hat{\H}$ is computed,
the $K$-means algorithm is applied to $\hat{\H}'$,
treating each row of $\hat{\H}$ as the spectral embedding of the associated $v_i$ vertex.
The next section, \sref{sec:selfexpressive}, now describes some subspace clustering methods that use this technique by first forming a weight matrix \W
and then applying the spectral clustering method on it.

\subsubsection{Self-Expressive Methods}
\label{sec:selfexpressive}

Self-expressive methods exploit
the ``self-expressiveness property'' \cite{sc_survey} of data
that hypothesizes that a single data point can be expressed
as a linear combination of other data points in its cluster,
which is trivially true if the data exactly follows a subspace model.
The goal is to learn those linear coefficients which is often achieved by adopting different regularizers in the formulation.
These self-expressive algorithms, e.g.,
SSC \cite{ssc} and LRSC \cite{lrsc}, 
learn the coefficient matrix $\Pb \in \mathbb{R}^{N \times N}$
by solving special cases of the following general optimization problem
\begin{equation}
\argmin_{\Pb} \text{DF}(\Y-\Y \Pb) + \lambda \Lambda(\Pb) \st \Pb \in \mathcal{S}_{\Pb}.
\label{eq:self_expressive}
\end{equation}
The function $\text{DF}(\Y- \Y \Pb)$ is a data fidelity term, $\Lambda(\Pb)$ is a regularizer,
and $\mathcal{S}_{\Pb}$ is some constrained set to encourage $\Pb$ to satisfy certain conditions.
In the case of SSC,
the optimization problem is formulated as
\begin{equation}
    \argmin_{\Pb} \normfrob{ \Y - \Y \Pb }^2 + \lambda \| \Pb \|_{1,1} \st \Pb_{ii} = 0
    \hspace{2mm} \forall i
    \label{eq:ssc}
\end{equation}
where $\| \Pb \|_{1,1} = \| \text{vec}(\Pb) \|_1 = \sum_i |p_i|$
is the vectorized 1-norm
and $\normfrob{ \cdot } = \| \text{vec}(\cdot) \|_2$ is the Frobenius norm of a matrix.
Observe that \eref{eq:ssc} approximates each data sample
as a sparse linear combination of other data points
to form $\Pb$.
Let $\text{abs}(\Pb)$ represent the element-wise absolute value of matrix $\Pb$. 
Then, spectral clustering is performed on
$\W = \frac{1}{2} (\text{abs}(\Pb') + \text{abs}(\Pb))$
by applying the $K$-means method
to the spectral embedding of the affinity matrix $\W$.
Ideally, \eref{eq:ssc} would select the high quality data to represent worse samples in $\Pb$
and this information would be retained in $\W$.
However, this data quality awareness condition is not guaranteed in self-expressive methods
to our knowledge. In our experiments, \fref{fig:heatmap_adssc} and \fref{fig:quasar_clustering} show that there must be an issue constructing an ideal $\Pb$ due to worse clustering performance in heteroscedastic conditions.

\subsubsection{Subspace Clustering via Thresholding (TSC)}

In the example to follow,
assume that there are two distinct 1-dimensional subspaces.
Given two data samples $\y_i$ and $\y_j$,
intuitively their dot product $\langle \y_i, \y_j \rangle$ will be high
if $c_i = c_j$, meaning they belong to the same subspace.
Likewise, if $c_i \neq c_j$,
then $\langle \y_i, \y_j \rangle = 0$
if the subspaces are orthogonal.
In non-orthogonal scenarios,
one expects
$|\langle \y_i, \y_j \rangle|$
to be smaller
when $c_i \neq c_j$
than when $c_i = c_j$. 
Using this idea, in more general D-dimensional subspace settings,
TSC constructs a matrix $\Z \in \mathbb{R}^{N \times N}$ that describes similarity such that
\begin{equation}
    \Z_{ij} = \exp(-2 \arccos(| \langle \y_i, \y_j \rangle | ))
    \st \hspace{2mm} \Z_{ii}=0 \hspace{1mm} \forall i.
\end{equation}
This matrix is then thresholded to retain only the top $q$ values for each row, i.e.,
\begin{equation}
    \W = \argmin_{\W} \normfrob{ \W - \Z }^2 \st \| \W_{i,:} \|_0 = q \hspace{2mm} \forall i
    \label{eq:threshold}
\end{equation}
where $\| \W_{i,:} \|_0 = q$ is the $l_0$ pseudo-norm that counts the number of nonzero entries. 
In other words, one constructs $\W_{i,:}$ using the q nearest neighbors
that correspond to the highest magnitude values.
Then, spectral clustering is applied to \W
to find the cluster associations. 

\subsubsection{Ensemble \texorpdfstring{$K$}{K}-Subspaces (EKSS)}

Iterative methods are based on the idea of alternating between cluster assignments
and subspace basis approximation,
with KSS being highly prominent \cite{kplane}. 
The KSS algorithm seeks to solve
\begin{equation}
\argmin_{\cC, \, \mathcal{U}} \sum_{k=1}^K
\sum_{
c_i = k, \forall i } \| \y_i - \U_k \U_k' \y_i \|_2^2 .
\label{eq:kss_model}
\end{equation}
The goal is to minimize the sum of residual norms
by alternating between performing PCA on each cluster to update $\mathcal{U}$ 
and using the subspace bases to calculate new cluster assignments \cC
in a similar fashion to the $K$-means algorithm. 
The quality of the solution depends highly on the initialization.
Recent work provides convergence guarantees and spectral initialization schemes
that provably perform better than random initialization \cite{kss_convergence}.
However, this problem \eref{eq:kss_model}, is known to be NP-hard \cite{gitlin2018improving}. Further, it is prone to local minima \cite{timmerman2013subspace}.
To overcome this, consensus clustering \cite{consensus_clustering}
is a tool that leverages information from many trials and combines results together.
This approach, known as Ensemble KSS (EKSS) \cite{ekss},
creates an affinity matrix
whose $(i,j)$th entry represents
the number of times the two points were clustered together in a trial.
Then, spectral clustering is performed on the affinity matrix
to get the final clustering from the many base clusterings.
The use of PCA makes it challenging to learn clusters and subspace bases
in the heteroscedastic regime
since PCA implicitly assumes the same noise variance across all samples
\cite{heppcat}.
The proposed ALPCAHUS approach in \sref{paper:method}
builds on KSS, and its ensemble version, by generalizing \eref{eq:kss_model} to the heteroscedastic regime while simultaneously learning data quality.

\subsection{Other heteroscedastic models}
\label{sec:other_models}

This paper focuses on heteroscedastic noise across the data samples.
There are other \emph{subspace learning} methods in the literature
that explore heteroscedasticity in different ways.
For example, HeteroPCA considers heteroscedasticity across the feature space
\cite{hetero_pca}.
One possible application of that model
is for data that consists of sensor information with multiple devices
that naturally have different levels of precision and signal to noise ratio.
Another heterogeneity model
considers the noise to be homoscedastic
and instead assumes that the signal itself is heteroscedastic \cite{collas2021probabilistic}.
That work considers applications
where the power fluctuating signals, i.e., heteroscedastic signals, are embedded in white Gaussian noise.
However, to the authors' knowledge, there are no algorithms for \emph{subspace clustering} that consider feature-space heteroscedasticity.
We exclude comparisons with these methods
since their models are very different and each have their own applications.

\subsection{Single Subspace Heteroscedastic PCA}
\label{sec:works_pca}

Before extending \eref{eq:kss_model} to the heteroscedastic setting,
we will review how LR-ALPCAH
\cite{alpcah_journal}
solves for a single subspace ($K=1$)
in the heteroscedastic setting
using the model described in \eref{eq:heteroscedastic_subspace_model}. 
For the measurement model
$\yi \sim \mathcal{N}(\x_i, \nu_i \I)$
in \eqref{eq:heteroscedastic_subspace_model},
the probability density function for a single data sample \yi following a Gaussian distribution is easily expressed as
\begin{equation}
 \frac{1}{\sqrt{(2\pi)^D |\nu_i \I |}}
 \exp{ [-\frac{1}{2} (\yi-\x_i)' (\nu_i \I)^{-1} (\yi-\x_i)] }.
 \label{eq:alpcah_nll1}
\end{equation}
For independent samples, after dropping constants, 
the joint log likelihood of all data $\{\yi\}_{i=1}^{N}$ is the following
\begin{equation}
    \sum_{i=1}^{N} -\frac{1}{2} \log |\nu_i \I|
    -\frac{1}{2} (\yi - \x_i)' (\nu_i \I)^{-1} (\yi - \x_i).
    \label{eq:alpcah_nll2}
\end{equation}
Let $\Pi = \mathrm{diag}(\nu_1,\ldots,\nu_N) \in \mathbb{R}^{N \times N}$
be a diagonal matrix representing the typically unknown noise variances.
Then, the negative log likelihood in matrix form is
\begin{equation}
\frac{D}{2} \log |\Pi| + \frac{1}{2} \text{Tr}[(\Y - \X) \Pi^{-1} (\Y - \X)'].
\label{eq:alpcah_nll3}
\end{equation}
After further manipulation by trace lemmas,
we rewrite the negative log-likelihood as
\begin{equation}
    \frac{1}{2} \normfrobr{ (\Y - \X) \Pi^{-1/2} }^2
    + \frac{D}{2} \log |\Pi|.
    \label{eq:alpcah_cost}
\end{equation}
Our previous work used a functional operator similar to the nuclear norm
to regularize $\X$ and encourage a low-rank solution
by penalizing the tail sum of singular values \cite{locallylowrank} \cite{alpcah}.
However, this method named ALPCAH was relatively slow due to SVD operations every iteration. 
To reduce computation and enforce a low-rank solution,
the LR-ALPCAH variant \cite{alpcah_journal} took inspiration
from the matrix factorization literature
\cite{matrix_factorization}
and factorized $\X \in \mathbb{R}^{D \times N} \approx \L \R'$
where $\L \in \mathbb{R}^{D \times \hat{d}}$
and $\R \in \mathbb{R}^{N \times \hat{d}}$
for some rank estimate $\hat{d}$.
Using this idea,
LR-ALPCAH
estimates \X
by solving for \L and \R, jointly with noise variance matrix $\Pi$, by the following
\begin{equation}
    \argmin_{\L, \, \R,\, \Pi}
    \frac{1}{2} \normfrobr{ (\Y - \L \R') \Pi^{-1/2} }^2 + \frac{D}{2} \log |\Pi|.
    \label{eq:lr_alpcah}
\end{equation}
This comes from a modified model of \eref{eq:heteroscedastic_subspace_model} described by
\begin{equation}
    \y_i = \L \r_i + \bepsilon_i , \quad \bepsilon_i \sim \mathcal{N}(\0, \nu_i \I)
    \label{eq:alpcah_nll4}
\end{equation}
where $\r_i$ denotes the $i$th column of $\R$. 
\sref{paper:method} combines ideas from \eref{eq:kss_model} and \eref{eq:lr_alpcah}
to tackle the union of subspaces setting.

\section{Proposed Subspace Clustering Method}
\label{paper:method}

For notational simplicity,
let 
$\Y_k = \Y_{\Ck} \in \mathbb{R}^{D \times N_k}$
denote the submatrix of \Y
having columns corresponding to data samples
that are estimated to belong in the $k$th subspace,
i.e., $\Y_k = \Y_{\Ck} = \text{matrix}( \{ \y_i \, : \, c_i = k \})$,
where
$\Ck = \{i \, : \, c_i = k\}$
must be determined.
This notation applies similarly to other matrices
such as
$\Pi_k = \Pi_{\Ck} = \text{diag}(\{\nu_i \, :\, c_i = k \})$.
For the union of subspace measurement model in 
\eref{eq:uos_heteroscedastic_model},
we generalize
LR-ALPCAH \eref{eq:lr_alpcah}
by proposing the following cost function
\begin{equation}
\argmin_{\mathcal{L}, \, \mathcal{R}, \, \Pi, \, \cC}
\underbrace{
\sum_{k=1}^K 
\overbrace{
\frac{1}{2} \normfrob{ (\Y_k - \L_k \R_k' ) \Pi^{-1/2}_{k} }^2 + \frac{D}{2} \log | \Pi_k | 
}^{ f_k(\L_k,\R_k,\Pi_k; \Y_k) } 
}_{ f(\mathcal{L}, \mathcal{R}, \Pi, \cC) }
\label{eq:alpcahus}
\end{equation}
where $\cC, \Pi, \mathcal{L}, \mathcal{R}$
denote the sets of estimated clusters, noise variances,
and factorized matrices respectively for each cluster $k = 1, \ldots, K$.
Specifically, $\mathcal{L} \defequ \{ \L_1, \ldots, \L_K \}$
and $\mathcal{R} \defequ \{ \R_1, \ldots, \R_K \}$.
Our algorithm for solving \eref{eq:alpcahus} is called
ALPCAHUS (\textbf{ALPCAH} for \textbf{U}nion of \textbf{S}ubspaces).
We solve \eqref{eq:alpcahus} via alternating minimization, and 
the method alternates between subspace basis estimation and data sample reassignment/clustering.
We begin with subspace basis estimation.

Firstly, with data sample assignments fixed, $\forall k \in \{1, \ldots,K\}$ we estimate subspace bases 
by applying $T_1$ iterations of alternating minimization
by
\begin{align}
\left( \L_k^{(T_1)}, \R_k^{(T_1)}, \Pi_k^{(T_1)} \right)
= \argmin_{\L_k, \, \R_k, \, \Pi_k}
&f_k (\L_k, \R_k, \Pi_k ; \Y_k) \nonumber \\
&\st \Pi_k \succeq \alpha \I.
\label{eq:lr-alpcah}
\end{align}
The solution to \eref{eq:lr-alpcah} is described in Ref.~\cite{alpcah_journal}.
However, for completeness and notational consistency,
the updates are included in
\eref{eq:lr_l_update}, \eref{eq:lr_r_update}, and \eref{eq:lr_pi_update}.
Given the current $t_1$ iteration, the updates for the $t_1 + 1$ iteration are given as
\begin{align}
    \L^{(t_1+1)}_k &= \argmin_{\L_k} f_k \left( \L_k, \R^{(t_1)}_k, \Pi^{(t_1)}_k; \Y_k^{(t_2)} \right)
    \nonumber \\
    &= \Y_k^{(t_2)} \Pi^{(t_1)}_k \R^{(t_1)}_k \left( \big(\R^{(t_1)}_k\big)' \Pi^{(t_1)}_k \R^{(t_1)}_k \right)^{-1}
    \label{eq:lr_l_update}\\
    \R^{(t_1+1)}_k &= \argmin_{\R_k} f_k \left( \L^{(t_1+1)}_k, \R_k, \Pi^{(t_1)}_k; \Y_k^{(t_2)} \right)
    \nonumber \\
    &= \big( \Y_k^{(t_2)} \big)'  \L^{(t_1+1)}_k \left( \big( \L_k^{(t_1+1)} \big)' \L^{(t_1+1)}_k \right)^{-1} .
    \label{eq:lr_r_update}
\end{align}
Further, estimate the noise variance terms by
\begin{equation}
    \underbrace{
    \max \left( \alpha, \frac{1}{|D|} \left\| \left( \Y_k^{(t_2)} - \L^{(t_1+1)}_k \big( \R_k^{(t_1+1)} \big)' \right) \e_i \right\|_2^2 \right)
    }_{ p_k(i) }
    \label{eq:z_equation}
\end{equation}
where $\e_i$ denotes the $i$th canonical basis vector
that is used to select the $i$th residual column and $\alpha>0$ is
a user-selected noise variance threshold parameter.
Then, we update the noise variance matrix $\Pi_k$
as follows
\begin{align}
    \Pi^{(t_1+1)}_k &= \argmin_{\Pi_k} f_k(\L^{(t_1+1)}_k, \R^{(t_1+1)}_k, \Pi_k; \Y_k^{(t_2)})
    \nonumber \\
    &= \text{diag}(p_k(1), \ldots,p_k(|\mathcal{C}_k|)).
    \label{eq:lr_pi_update}
\end{align}
Because $\Pi_k$ is a diagonal matrix,
the $\Pi_k \succeq \alpha \I$ majorization condition in \eref{eq:lr-alpcah}
being equivalent to $\forall i,  \lambda_i (\Pi_k) \geq \alpha$
implies that $\forall i, \nu_i > \alpha$.
This leads to a projection to the positive set $[\alpha, \infty)$ by 
the $\max(\alpha, \cdot)$ condition in \eref{eq:z_equation}.
This condition is sufficient to ensure convergence as proven
in \cite[Thm. 2]{alpcah_journal}
and used in this work to further prove ALPCAHUS convergence
in \tref{thm:alpcahus} using some $\alpha \in \mathbb{R}^{+}$.

\begin{algorithm*}
\renewcommand{\thealgorithm}{ALPCAHUS}
    \caption{(unknown noise variances, unknown noise grouping, ensemble version with random init.)}
    \label{alg:alpcahus}
    \begin{algorithmic}[1]
        \STATE \textbf{Input:} $ \Y \in \mathbb{R}^{D \times N}$: data,
        $K \in \mathbb{Z}^{+}$: number of subspaces,
        $ \{\hat{d}_k \in \mathbb{Z}^{+} \hspace{2mm} \forall k \in \{1,\ldots,K\} \}$:
        candidate dimension for all clusters,
        $q \in \mathbb{Z}^{+}$: threshold parameter,
        $B \in \mathbb{Z}^{+}$: base clusterings,
        $T_1 \in \mathbb{Z}^{+}$: LR-ALPCAH iterations,
        $T_2 \in \mathbb{Z}^{+}$: maximum alternating updates (cluster reassignments)
        \STATE \textbf{Output:} $\cC_S = \{c_1, \dots,c_N\}$: clusters of $\Y$
        \FOR{$b = 1,\dots,B$ \green{(in parallel)}}
        
        \STATE $\Ck \sim \{1,\ldots,N\} \st |\Ck| \approx \frac{N}{\Bar{K}}$ for $k = 1,\dots,K$ 
        \hfill \green{Initialize clusters randomly}
        \STATE $\Pi_k \gets \I$ for $k = 1,\dots,K$ \hfill \green{Assume homoscedastic data initially}
        \vspace{1mm}
        \STATE $\L_k, \R_k \gets
        \U_{:,1:\hat{d}_k} \Sig^{1/2}_{1:\hat{d}_k,1:\hat{d}_k}, \
        \Sig^{1/2}_{1:\hat{d}_k,1:\hat{d}_k} \V_{:,1:\hat{d}_k}
        \text{ where }
        \text{SVD}(\Y_{\mathcal{C}_k})
        = \U \Sig \V'
        $
        for $k = 1,\dots,K$
        \hfill \green{Spectral init}
        \vspace{1mm}
        
             \FOR{$t_2 = 1,\dots,T_2$ \green{(in sequence)}}
             \STATE $\L_{k}, \R_{k}, \Pi_{k} \gets$ 
             \text{Compute} \eref{eq:lr_l_update}-\eref{eq:lr_pi_update}
             \text{using} $\Y_{\mathcal{C}_k}$ for $T_1$ iterations for $k = 1,\dots,K$
             \hfill \green{Apply LR-ALPCAH \cite{alpcah_journal}}

            \STATE  $\Ck \gets \{\forall \y_i \in \Y : \forall k, $
            \text{apply} \eref{eq:cluster_cost}-\eref{eq:cluster_update}
            with $\U_k \text{ extracted from GramSchmidt}(\L_k)\}$
            \hfill \green{Update cluster labels}
            \STATE Stop execution if $\mathcal{C}_k^{(t_2+1)} = \mathcal{C}_k^{(t_2)}$
            is met for all clusters \label{alg:finite_iterations} \hfill \green{Stopping criteria}
             \ENDFOR
             \STATE $\cC^{(b)} \gets \mathcal{C} = \{c_{1},\dots,c_{N} \} $
             \hfill \green{Collect results from all trials}
        \ENDFOR
        \STATE $\A_{i,j} \gets$ Form $\A$ by \eref{eq:coassociation_matrix} using
        $\{ \mathcal{C}^{(1)}, \ldots, \mathcal{C}^{(B)}\}$
        \hfill \green{Form affinity matrix of similar clusterings}

        \STATE $\Z^\text{row} \gets$ Threshold rows by solving \eref{eq:threshold_rows} using $\A$
        \hfill \green{Retain top $q$ elements for each row}
        \STATE $\Z^\text{col}\ \gets$ Threshold columns by solving \eref{eq:threshold_columns} using $\A$
        \hfill \green{Retain top $q$ elements for each column}
  
        \STATE $\W \gets \frac{1}{2}\paren{\Z^\text{row}+\Z^\text{col}}$
        \hfill \green{Average affinity matrix}
        \STATE $\cC_S  \gets$ Perform spectral clustering on $\W$ via \eref{eq:ratio_cut}-\eref{eq:spectral_clustering} 
        \hfill \green{Final clustering}
    \end{algorithmic}
\end{algorithm*}

Secondly, fixing the subspaces, we update the data sample assignments to clusters.
The cluster update is essentially
\(
\cC\supnew = \argmin_{\cC} f(\cL, \cR, \Pi, \cC)
,\)
with a rule
to break ties in favor
of the previous cluster assignment
as follows.
Let $\U_k^{(T_1)}$ denote the Gram-Schmidt vectors of $\L_k^{(T_1)}$.
Define residual point error for $k$th basis by
\begin{equation}
J_{i}(k) = \left\| \y_{i} -  \U_k^{(T_1)} \left( \U_k^{(T_1)} \right)' \y_{i} \right\|_2^2
\label{eq:cluster_cost}
\end{equation}
and let the set $S_{J_i}$ denote the set of minimizers of $J_i(k)$ by
\begin{equation}
\mathcal{S}_{J_i} = \argmin_k J_i(k).
\label{eq:min_cluster_set}
\end{equation}
Then $\forall i$,
given the label estimate $c_i^{(t_2)}$ for the $t_2$ iteration,  compute the next label by
\begin{align}
c_i^{(t_2+1)} = \left\{
\begin{array}{ll}
c_i^{(t_2)} & \text{ if } c_i^{(t_2)} \in \mathcal{S}_{J_i} \\
k^{*} \in \mathcal{S}_{J_i} & \text{ otherwise}.
\end{array}
\right.
\label{eq:cluster_update}
\end{align}
The solution to \eref{eq:cluster_update} involves the following procedure.
Given a data sample $\y_i$,
find the lowest subspace projection residual out of all subspaces by \eref{eq:cluster_cost} 
and assign it to $c_i^{(t_2+1)}$ at the $t_2+1$ iteration.
In the event of a tie,
meaning there is more than one subspace that has equal residual,
retain the past label $c_i^{(t_2)}$ for the new label $c_i^{(t_2+1)}$.
By doing this, cycling between labels is prevented for all data samples.
This cluster reassignment criteria will be important to ensure convergence as proven in \tref{thm:alpcahus}.
Because \eref{eq:min_cluster_set} is with respect to $k$
and not the $i$th sample,
a residual weighted with
the noise parameter $\nu_i$ is unnecessary, since it would not change the minimizing $k$. 

To further improve clustering performance,
consensus clustering can be leveraged over many trials.
Initially, we tried using this approach with only one trial
as seen in \fref{fig:heatmap_alpcah}
and ALPCAHUS ($B=1$) result in \fref{fig:astro_cluster}.
However, since $K$-subspaces in general is sensitive to initialization,
during experimentation we found higher clustering accuracy by using a consensus approach with more than one trial
as shown in \fref{fig:heatmap_alpcahus}
and ALPCAHUS ($B=16$) result in \fref{fig:astro_cluster}.

\subsection{Ensemble Extension for ALPCAHUS}
\label{sec:method_ensemble}
This section presents the ensemble algorithm with base clustering parameter $B$
to combine multiple trials for finding \cC in \eref{eq:alpcahus}.
ALPCAHUS with $B>1$ leverages consensus clustering
by forming an affinity matrix $\A \in \mathbb{R}^{N \times N}$ where
\begin{align}
    \A_{i,j} = &\frac{1}{B} | \{\forall b \in \{1,\ldots,B\} \text{ such that } \nonumber \\
    &\y_i, \y_j \text{ are co-clustered in } \cC^{(b)} \} |
    \label{eq:coassociation_matrix}
\end{align}
and $\cC^{(b)} = \{ c_1^{(b)}, \ldots, c_N^{(b)} \}$
refers to the cluster labels for the $b$th trial ranging from $1$ to $B$.
Then, the rows and columns of $\A$ are thresholded to retain the top $q$ values by solving
\begin{align}
    \Z^{\text{row}} = \argmin_{\Z} \normfrob{ \A - \Z }^2
    \st \| \Z_{i,:} \|_0 = q \hspace{2mm} \forall i \label{eq:threshold_rows} \\
    \Z^{\text{col}} = \argmin_{\Z} \normfrob{ \A - \Z }^2
    \st \| \Z_{:,i} \|_0 = q \hspace{2mm} \forall i. \label{eq:threshold_columns}
\end{align}
Finally, spectral clustering is applied to
$\W = \frac{1}{2}(\Z^{\text{col}} + \Z^{\text{row}})$
to get the clusters from the results of the ensemble.
This spectral clustering operation assumes a balanced cluster distribution
that may not hold in practice.
All of our included experiments contain balanced cluster sizes;
however, \sref{appendix:balancedcluster} provides
discussion of this limitation and possible remediation strategies.

The $q$ threshold parameter can be set by cross validation, which is what is done in the experiments shown in \sref{paper:experiments}, or other techniques can be used. For example, Ref. \cite{l2graph_subspace_clustering} creates a sparse L2 graph through hard thresholding to retain the top $q=d_k$ values where $d_k$ is the known subspace dimension. Thus, $q$ can be tied to the true or estimated subspace dimension instead.
\sref{appendix:parameters} provides more discussion on ALPCAHUS hyperparameters including recommended values.

In this more general ensemble framework, one could select $B=1$
to reduce to one trial of the optimization problem in \eref{eq:alpcahus} which is the non-ensemble version of ALPCAHUS.
\aref{alg:alpcahus}
summarizes the procedure for subpace clustering with optional ensemble learning. For Julia code implementations, refer to \url{https://github.com/javiersc1/ALPCAHUS}.

\subsection{ALPCAHUS Convergence}
In the single cluster setting,
ALPCAHUS with $K=1$ has a sequence of cost function values
$f_1(\L_1, \R_1, \Pi_1; \Y_1)$
that provably converges to local minima,
since the cost function and algorithm simplify to LR-ALPCAH \eref{eq:lr_alpcah} 
that has convergence guarantees as proven in Ref.~\cite[Thm. 2]{alpcah_journal}.
In the multi-cluster setting with $K > 1$, \tref{thm:alpcahus} below
shows that
the sequence of cost function values produced
by $f(\mathcal{L}, \mathcal{R}, \Pi, \cC)$ in \eref{eq:alpcahus}
converges asympototically.
The argument in \tref{thm:alpcahus} below
is a natural extension of that in the original k-subspaces paper \cite[Thm. 7]{kplane}
with greater mathematical expos\'e.
\sref{appendix:convergence} provides
an experimental result that corroborates this theorem.

\begin{theorem}
\label{thm:alpcahus}
Consider the ALPCAHUS cost function $f(\mathcal{L}, \mathcal{R}, \Pi, \cC)$ in \eref{eq:alpcahus}.
Assume a noise variance threshold parameter $\alpha \in \mathbb{R} > 0$
that lower bounds all $\nu_i$, 
and the cluster assignment criteria in \eref{eq:cluster_update} that accepts changes
only if there is a cluster reassignment of points
that strictly decreases the cost function $f(\cdot)$, as expressed in \eref{eq:cluster_update}.
Then, \eref{eq:alpcahus} generates a sequence of cost function values that converges asymptotically.
Furthermore, the algorithm terminates in finite iterations
due to stopping criteria in line \eqref{alg:finite_iterations} of \aref{alg:alpcahus}
that checks whether clusters have changed from the previous iteration.

\end{theorem}
\begin{proof}
To prove that the sequence of cost function values converges,
we show that each step decreases the cost function.
We also show that
the algorithm terminates at a point where the variables stop changing.
Each step consists of two sub-steps:
first, the subspace estimation sub-step where LR-ALPCAH is used,
and second, the clustering sub-step
where projection onto the estimated subspaces is done.
Let $t_1$ be the iteration variable associated with the variables in the subspace estimation sub-step
and $t_2$ be the iteration variable associated with the variables in the clustering sub-step. 
Define $\X_k =\L_k \R_k' $
and bound the cost by
\begin{align}
&f(\mathcal{L}^{(t_1)}, \mathcal{R}^{(t_1)}, \Pi^{(t_1)}, \cC^{(t_2)}) = \nonumber \\
&\frac{1}{2} \sum_k \sum_{i \in \mathcal{C}_k^{(t_2)}} \frac{1}{\nu_i^{(t_1)}} \left\| [\Y_k]_i - [\X_k]_i^{(t_1)} \right\|_2^2 +D \log{ \nu_i^{(t_1)}} \label{step:sub0}  \\
&\geq \frac{1}{2} \sum_k \sum_{i \in \mathcal{C}_k^{(t_2)}} \frac{1}{\nu_i^{(t_1+1)}} \left\|[\Y_k]_i - [\X_k]_i^{(t_1+1)} \right\|_2^2 \nonumber \label{step:sub} \\
&+ D \log \nu_i^{(t_1+1)}  = f(\mathcal{L}^{(t_1+1)}, \mathcal{R}^{(t_1+1)}, \Pi^{(t_1+1)}, \cC^{(t_2)}) \\
    &\geq \frac{1}{2} \sum_k \sum_{i \in \mathcal{C}_k^{(t_2+1)}} \frac{1}{\nu_i^{(t_1+1)}}
    \left\| [\Y_k]_i - [\X_k]_i^{(t_1+1)} \right\|_2^2
    \nonumber \\
&+ D \log \nu_i^{(t_1+1)} = f(\mathcal{L}^{(t_1+1)}, \mathcal{R}^{(t_1+1)}, \Pi^{(t_1+1)}, \cC^{(t_2+1)}).
    \label{step:clust}
\end{align}
Here, \eqref{step:sub0} $\geq$ \eqref{step:sub} is due to the subspace step not increasing the cost for each $f_k$ term 
as shown in Ref. \cite[Thm. 2]{alpcah_journal},
and \eqref{step:sub} $\geq$ \eqref{step:clust} is due to 
the cluster reassignment criteria in \eref{eq:cluster_update} only updating the cluster label that decreases \eref{eq:cluster_cost}.
What has been shown is that
\begin{align}
    f^{(t_1,t_2)} \geq f^{(t_1+1,t_2)} \geq f^{(t_1+1,t_2+1)}
\end{align}
where $f^{(t_1,t_2)} = f(\mathcal{L}^{(t_1)}, \mathcal{R}^{(t_1)}, \Pi^{(t_1)}, \cC^{(t_2)})$. 
Thus,
as the cluster assignments and subspace bases are updated,
the cost is non-increasing. 
Since the cost function values are non-increasing at every step, bounded below,
then the sequence of cost function values produced by \eref{eq:alpcahus} converges asymptotically.
Furthermore, the stopping criteria in line \eqref{alg:finite_iterations}
ensures that the algorithm terminates
when all clusters do not change.
Because there are only a finite number of ways to assign cluster labels,
\aref{alg:alpcahus} will stop after a finite number of iterations.
\end{proof}

\subsection{Rank Estimation}
\label{sec:method_rank}
In subspace clustering,
some algorithms like SSC do not require subspace dimension to be known or estimated,
whereas others like KSS require it.
For the group of algorithms that require this parameter,
there is great interest in adaptive methods that can learn the subspace dimension.
In recent work, Ref. \cite{kss_convergence} proposed using an eigengap heuristic
on the sample covariance matrix of each cluster to estimate dimension.
Mathematically, this means calculating the following
\begin{align}
    \Sb_k &= \frac{1}{|\Ck|} \Y_k \Y_k' \\
    \hat{d}_k &= \argmax_{i} |\lambda_{i}(\Sb_k) - \lambda_{i+1}(\Sb_k)|
\end{align}
where $\lambda_i(S_k)$ denotes the $i$th eigenvalue of $S_k$
assuming $\lambda_1 \geq \ldots \geq \lambda_D$.
For heteroscedastic data, the eigengap heuristic can break down,
making it challenging to determine rank,
as shown in \sref{paper:experiments}, \fref{fig:rank_estimation}.
In recent work, Ref. \cite{signFlips} developed a parallel analysis algorithm
to estimate the rank of a matrix
that is consistently shown to work well in the heteroscedastic regime
if the data comes from a \emph{single} subspace only.
It works by creating an i.i.d. Bernoulli $(p=0.5)$ matrix
denoted as $\M$ and analyzing the singular values of
$\M \odot \Y$ for a matrix of data samples $\Y$.
One can distinguish what singular values are associated
with the signal and noise component of the data through this process.
We generalize this line of work and apply it to the union of subspace setting
by starting off over-parameterized and adaptively shrinking the bases by
\begin{align}
\forall k &\in \{ 1, \ldots , K\} \hspace{2mm} \text{do} \nonumber \\
\Tilde{\sigma}^{(r)} &= \text{SingularValues}(\M \odot \Y_k) \hspace{2mm}
\forall r \in \{1,\ldots,R\} \\
\Hat{d}_k &= \text{smallest } d \text{ that satisfies } \nonumber \\
&\sigma_{d+1} (\Y_k) \leq \alpha
\text{-percentile of } \{ \Tilde{\sigma}_{d+1}^{(1)}, \ldots, \Tilde{\sigma}_{d+1}^{(R)} \}.
\end{align}
This is done for each estimated cluster $\Y_k$
over $R$ random trials where
$1 \leq R \ll T_2$, and $T_2$ denotes the maximum ALPCAHUS iterations.
Here, $\sigma_{d+1} (\Y_k)$ denotes the $d+1$ singular value of $\Y_k$.
This process is repeated for all cluster subsets $\{ \Y_1, \ldots, \Y_K \}$
after the cluster reassignment update in \eref{eq:cluster_update}.
To reduce computation,
this dimension estimation is performed sparingly in the ALPCAHUS method.

\subsection{Cluster Initialization}
\label{sec:method_init}
For the non-ensemble version of ALPCAHUS ($B=1$), 
it previously remained to be seen
whether there exists an initialization scheme
that performs better than random cluster assignment in the heteroscedastic regime.
In recent work, Ref. \cite{kss_convergence} proposed
a thresholding inner-product based spectral initialization method (TIPS),
designed for homoscedastic data to be used with KSS,
where an affinity matrix is generated by $\W_{ij} = 1 \text{ if } | \langle \y_i , \y_j \rangle | \geq \tau$ and $i \neq j$ given a thresholding parameter $\tau > 0$.
Instead, in this work, we generate a fully connected $\W$ by creating
\begin{equation}
    \W_{ij} = |\langle \y_i, \y_j \rangle| \text{ if } i \neq j, \hspace{1mm} 0 \text{ otherwise},
    \label{eq:aij-tips}
\end{equation}
and hard threshold to retain the top $q$ edges by solving \eref{eq:threshold} so that $\tau=q$ for simplicity.
Then, the cluster assignments $\cC=\{c_1,\ldots,c_N\}$
are calculated by applying the spectral clustering method on the thresholded $\W$. Recall that $\y_i = \x_i +\bepsilon_i$.
Upon closer analysis, this metric in expectation, ignoring absolute value, gives
\begin{align}
\mathbb{E}[  \langle \y_i , \y_j \rangle  ]  &=
\mathbb{E}[ \langle \x_i,\x_j \rangle] + \mathbb{E}[ \langle \x_i,\bepsilon_j \rangle] \nonumber
\\ &+ \mathbb{E}[ \langle \bepsilon_i,\x_j \rangle]
+ \mathbb{E}[ \langle \bepsilon_i,\bepsilon_j \rangle ]
= \mathbb{E}[ \langle \x_i, \x_j \rangle ].
\label{eq:tips_metric}
\end{align}
Therefore, the metric is independent of the noise variances,
meaning the affinity matrix constructed does not have
highly unbalanced, asymmetric edge weights from noisy samples.
Thus this metric is more robust to heteroscedastic noise than others
such as Euclidean norm where
\begin{equation}
  \mathbb{E}[ \| \y_i - \y_j \|_2^2] = \| \x_i - \x_j \|_2^2 + D (\bnu_i + \bnu_j).  
\end{equation}
Clearly, this is inflated by the noise terms $\bnu_i$ and $\bnu_j$.
To note, TIPS initialization is only useful when $B=1$ for one base clustering
since it is a deterministic initialization;
it provably performs better than random initialization
as illustrated in \fref{fig:spectral_init}.
Otherwise, random initialization is used when $B>1$
to leverage consensus information in the ensemble process.

\section{Experiments}
\label{paper:experiments}

\subsection{Synthetic Experiments}

\subsubsection{Experimental Setup}

A synthetic dataset is generated consisting of $K=2$ clusters each of dimension $d = 3$
derived from random subspaces in $D=100$ dimensional ambient space.
Each cluster consists of two data groups
with group 1 containing $N_1 = 6$ samples, to explore the data constrained regime, with noise variance $\nu_1 = 0.1$ per cluster.
For group 2, the $N_2$ samples and noise variance $\nu_2$ are varied.
Cross validation is done for hyperparameters in any algorithms that require it with a separate training set. 
For a more detailed description of ALPCAHUS parameters, including recommended default values,
see \sref{appendix:parameters}. 
Both the estimated clusters from each clustering algorithm
and the known clusters are necessary to calculate clustering error. Further, the problem of label permutations is dealt with by using the Hungarian algorithm \cite{hungarian}. More concretely, clustering error (\%) is defined as
\begin{equation}
    \text{clustering error} = \frac{100}{N} \left( 1 - \max_{\pi} \sum_{i,j} \Q_{\pi(ij)}^{\text{out}} \Q_{ij}^{\text{true}} \right)
\end{equation}
where $\pi$ is a permutation of cluster labels found by applying the Hungarian algorithm,
and $\Q^{\text{out}}$ and $\Q^{\text{true}}$
are output labels and ground truth labelings of the data
where the $(i,j)$th entry is one if point $j$ belongs to cluster $i$ and zero otherwise.

The average clustering error is computed out of 100 trials where each trial had different noise, basis coefficients, and subspace basis realizations for the clusters.
Various subspace clustering algorithms are included such as $K$-Subspaces (KSS) \cite{k-subspaces}, Ensemble $K$-Subspaces (EKSS) \cite{ekss},
Subspace Clustering via Thresholding (TSC) \cite{heckel2015robust},
and Doubly Stochastic Sparse Subspace Clustering (ADSSC) \cite{adssc} which is a modern variant of SSC.
Applying general non-subspace clustering methods
such as $K$-means \cite{kmeans} resulted in poor clustering error so results of these methods are not shown in general. 

\begin{figure*}
\centering
\subcaptionbox{KSS (TIPS) mean clustering error. \label{fig:heatmap_kss}}{%
\includegraphics[width=0.4\columnwidth]{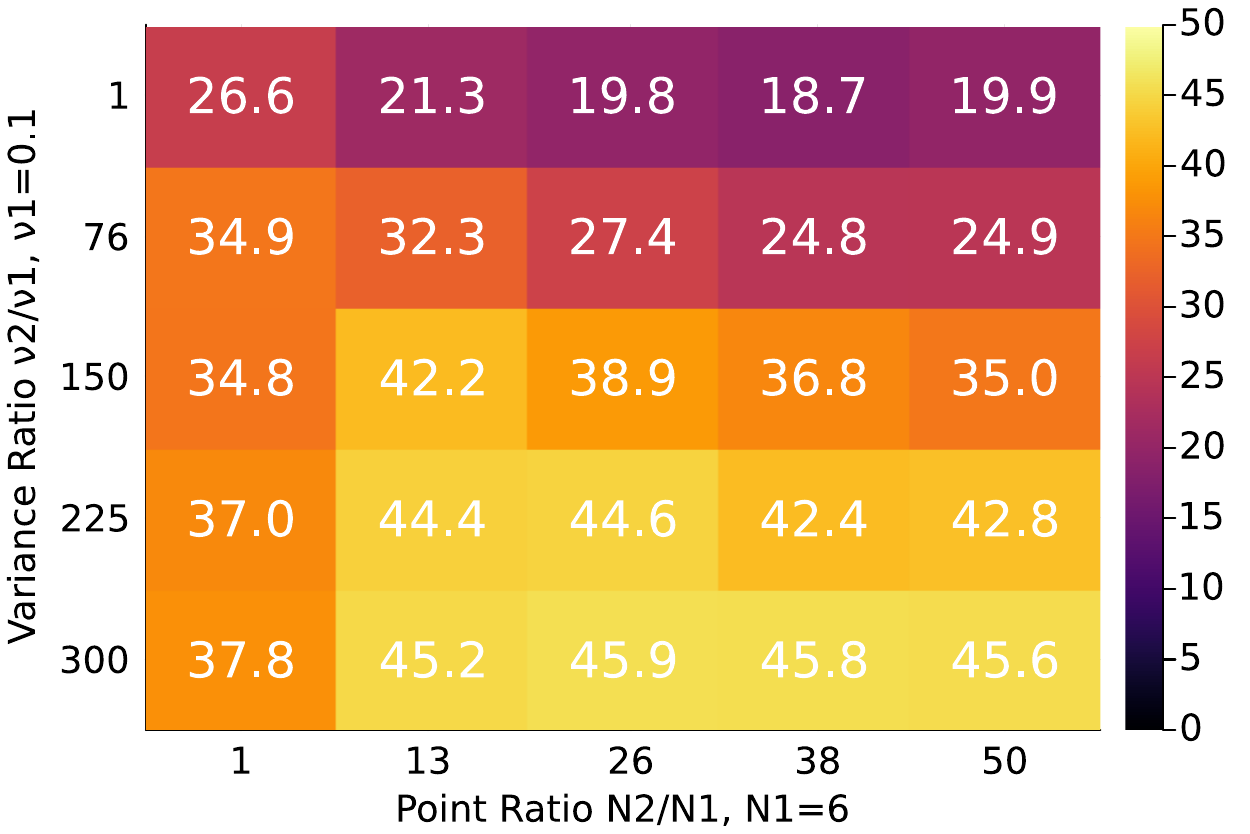}
}
\hfil
\subcaptionbox{ALPCAHUS ($B=1$, TIPS) mean clustering error. \label{fig:heatmap_alpcah}}{%
\includegraphics[width=0.4\columnwidth]{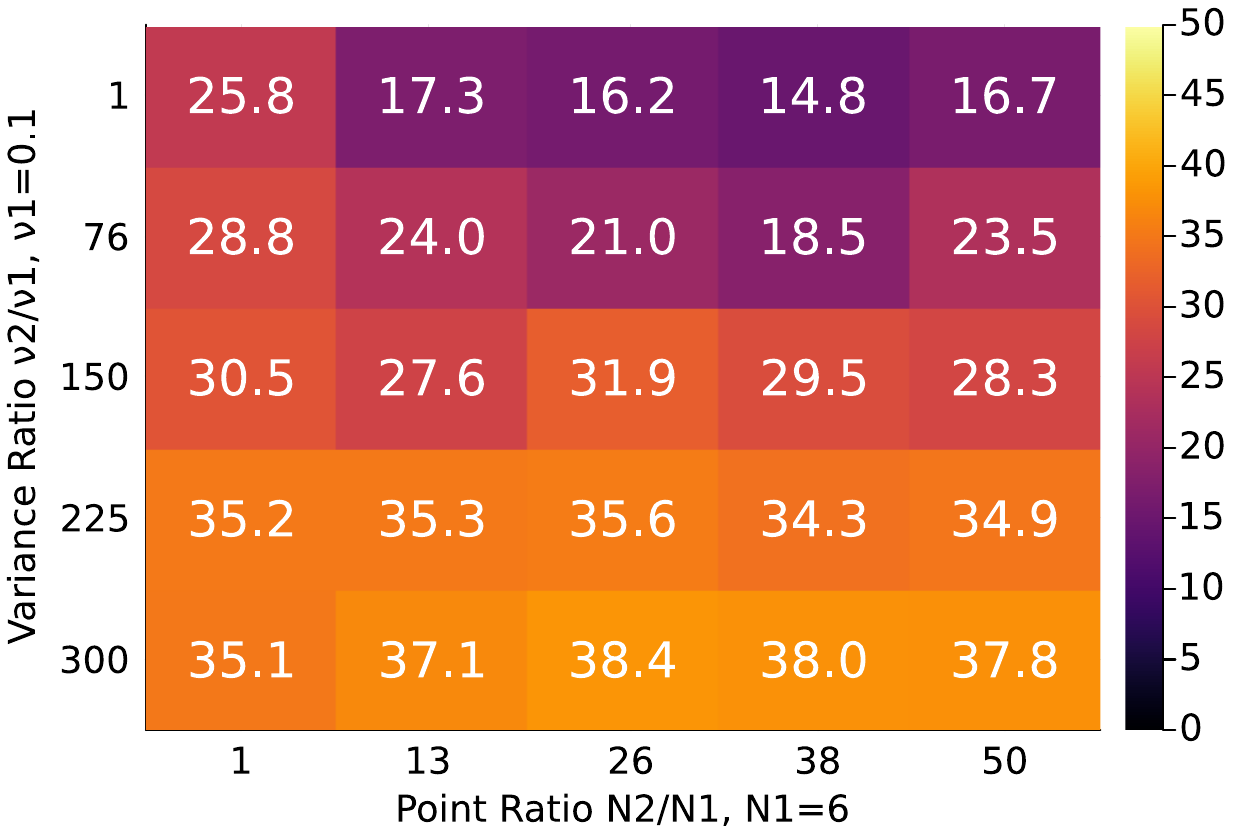}
}

\bigskip

\subcaptionbox{EKSS (B=128) mean clustering error. \label{fig:heatmap_ekss}}{%
\includegraphics[width=0.4\columnwidth]{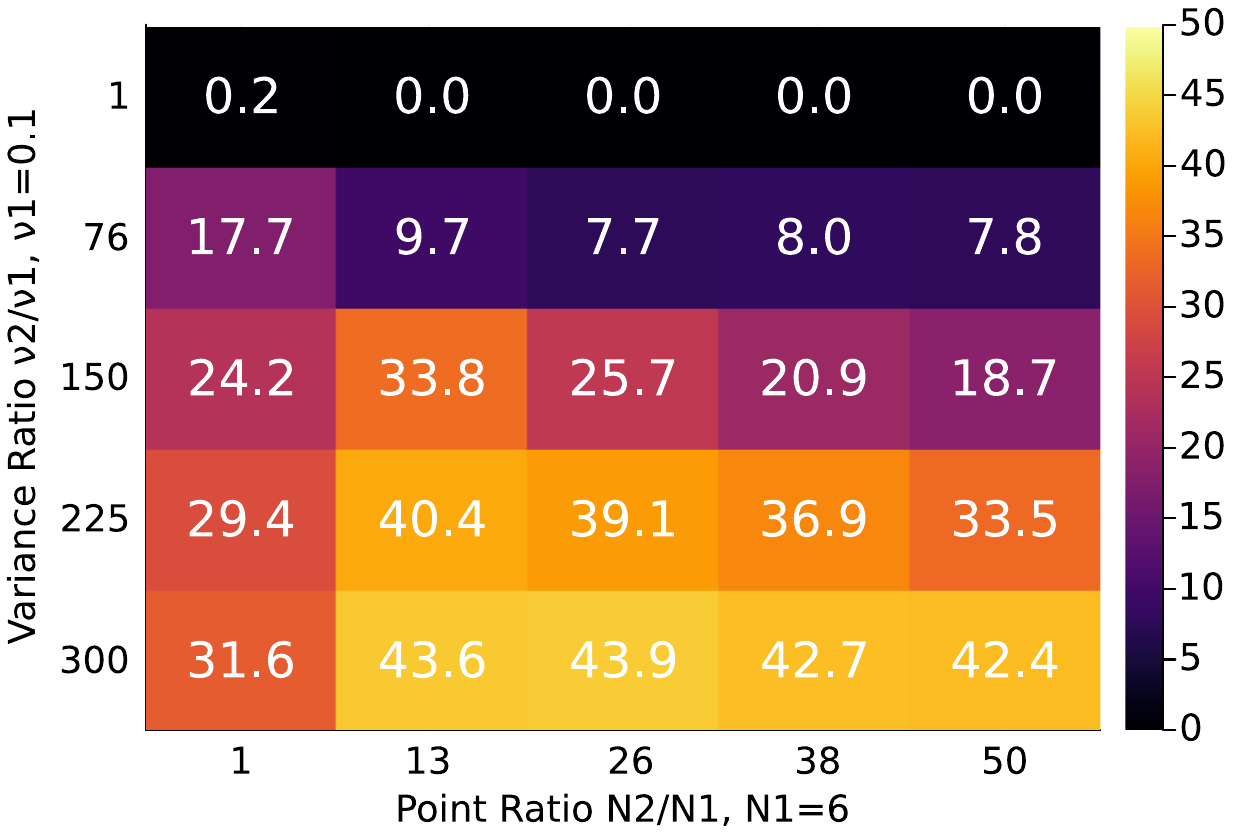}
}
\hfil
\subcaptionbox{ALPCAHUS (B=128) mean clustering error. \label{fig:heatmap_alpcahus}}{%
\includegraphics[width=0.4\columnwidth]{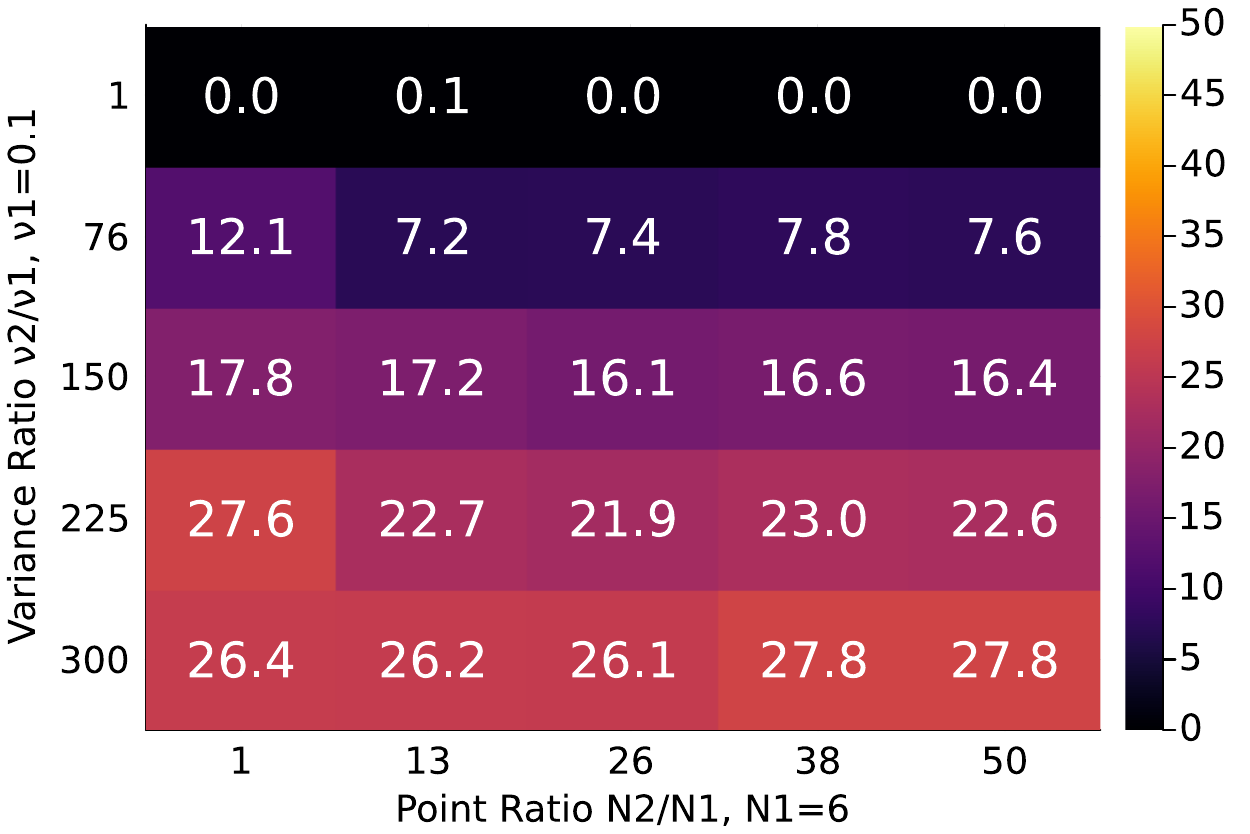}
}

\bigskip

\subcaptionbox{TSC mean clustering error. \label{fig:heatmap_tsc}}{%
\includegraphics[width=0.4\columnwidth]{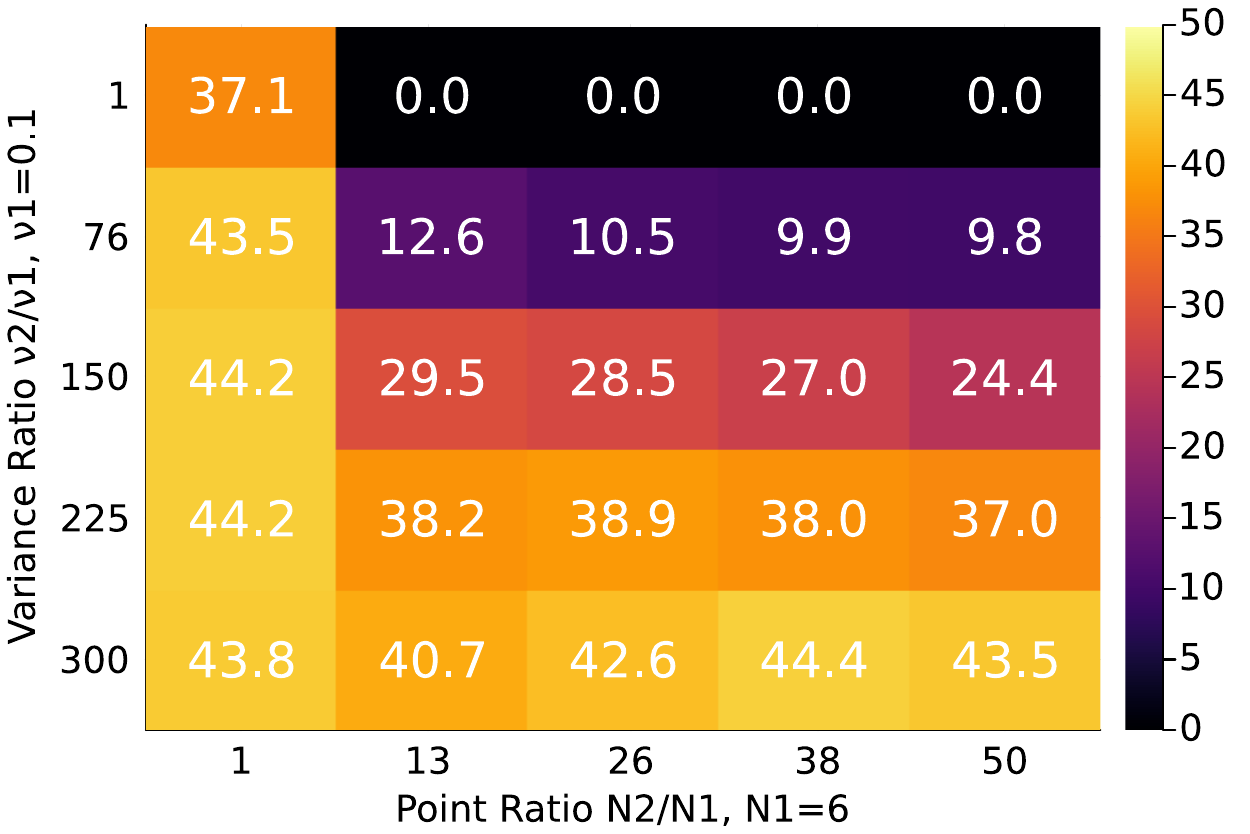}
}
\hfil
\subcaptionbox{ADSSC mean clustering error. \label{fig:heatmap_adssc}}{%
\includegraphics[width=0.4\columnwidth]{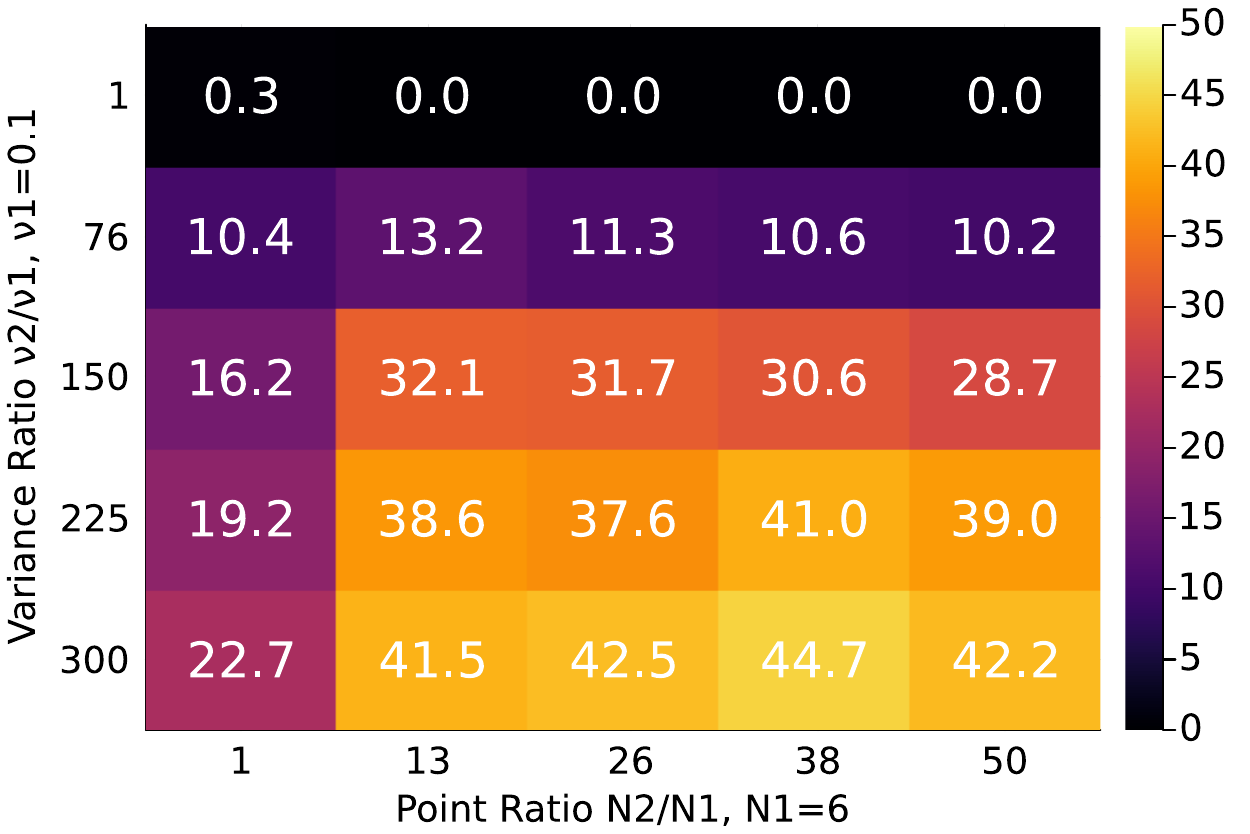}
}
\subcaptionbox{Noisy oracle mean clustering error. \label{fig:heatmap_oracle}}{%
\includegraphics[width=0.4\columnwidth]{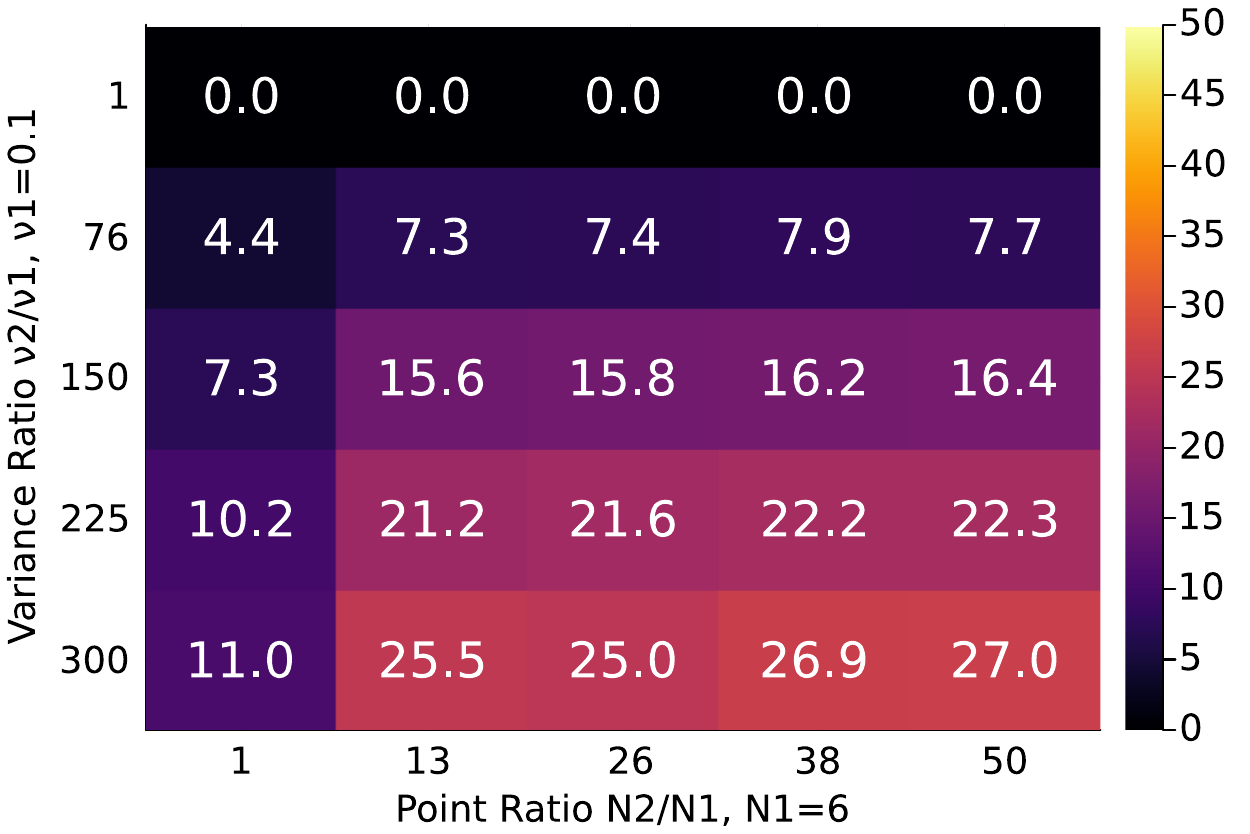}
}
\hfil
\subcaptionbox{Clustering error (\%) results for heteroscedastic landscape. Visual results included in \fref{fig:clustering_heatmaps}. Columns consist of the 4 corners and the 3 diagonal elements of the heatmaps in \fref{fig:clustering_heatmaps}. \label{fig:tab_clustering_landscape}}{

\resizebox{0.4\columnwidth}{!}{%
\begin{tabular}{|c|r|r|r|r|r|r|r|}
\hline
\multicolumn{1}{|c|}{$\nu_2/\nu_1$} & \multicolumn{1}{c|}{$1$} & \multicolumn{1}{c|}{$1$} &
\multicolumn{1}{c|}{$300$} & \multicolumn{1}{c|}{$300$} &
\multicolumn{1}{c|}{$150 $} & \multicolumn{1}{c|}{$225$} & \multicolumn{1}{c|}{$76$}
\\ \hline
\multicolumn{1}{|c|}{$N_2/N_1$} & \multicolumn{1}{c|}{$1$} & \multicolumn{1}{c|}{$ 50$} &
\multicolumn{1}{c|}{$1$} & \multicolumn{1}{c|}{$50$} & \multicolumn{1}{c|}{$26$} &
\multicolumn{1}{c|}{$13$} & \multicolumn{1}{c|}{$38$}
\\ \hline
\multicolumn{8}{|c|}{Reference Level} \\ \hline
{Noisy Oracle} & {0.0} & {0.0} & {11.0} & {27.0} & 15.8 & 21.2 & 7.9 \\ \hline
\multicolumn{8}{|c|}{Method Comparisons} \\ \hline
{\begin{tabular}[l]{@{}l@{}} \makecell{KSS} \end{tabular}}
& 26.6 & 19.9 & 37.8 & 45.6 & 38.9  & 44.4 & 24.8
\\ \hline
{\begin{tabular}[l]{@{}l@{}}
\makecell{EKSS \\ ($B=128$)}\end{tabular}} & 0.2 & \textbf{0.0} & 31.6 & 42.4 & 25.7 & 40.4 & 8.0 \\ \hline
\makecell{ADSSC} & 0.3 & \textbf{0.0} & \textbf{22.7} & 42.2 & 31.7 & 38.6 & 10.6 \\ \hline
\makecell{TSC} & 37.1 & \textbf{0.0} & 43.8 & 43.5 & 28.5 & 38.2 & 9.9 \\ \hline
{\begin{tabular}[l]{@{}l@{}} \makecell{ALPCAHUS \\ ($B=1$)} \end{tabular}}
& 25.8 & 16.7 & 35.1 & 37.8 & 31.9 & 35.3 & 18.5
\\ \hline
{\begin{tabular}[l]{@{}l@{}}
\makecell{ALPCAHUS \\ ($B=128$)}\end{tabular}}
& \textbf{0.0} & \textbf{0.0} & 26.4 & \textbf{27.8} & \textbf{16.1} & \textbf{22.7} & \textbf{7.8}
\\ \hline
\end{tabular}%
}

}

\caption{Clustering error over the heteroscedastic landscape
for various subspace clustering algorithms.}
\label{fig:clustering_heatmaps}
\end{figure*}

\begin{figure}
    \centering
    \includegraphics[width=0.8\textwidth]{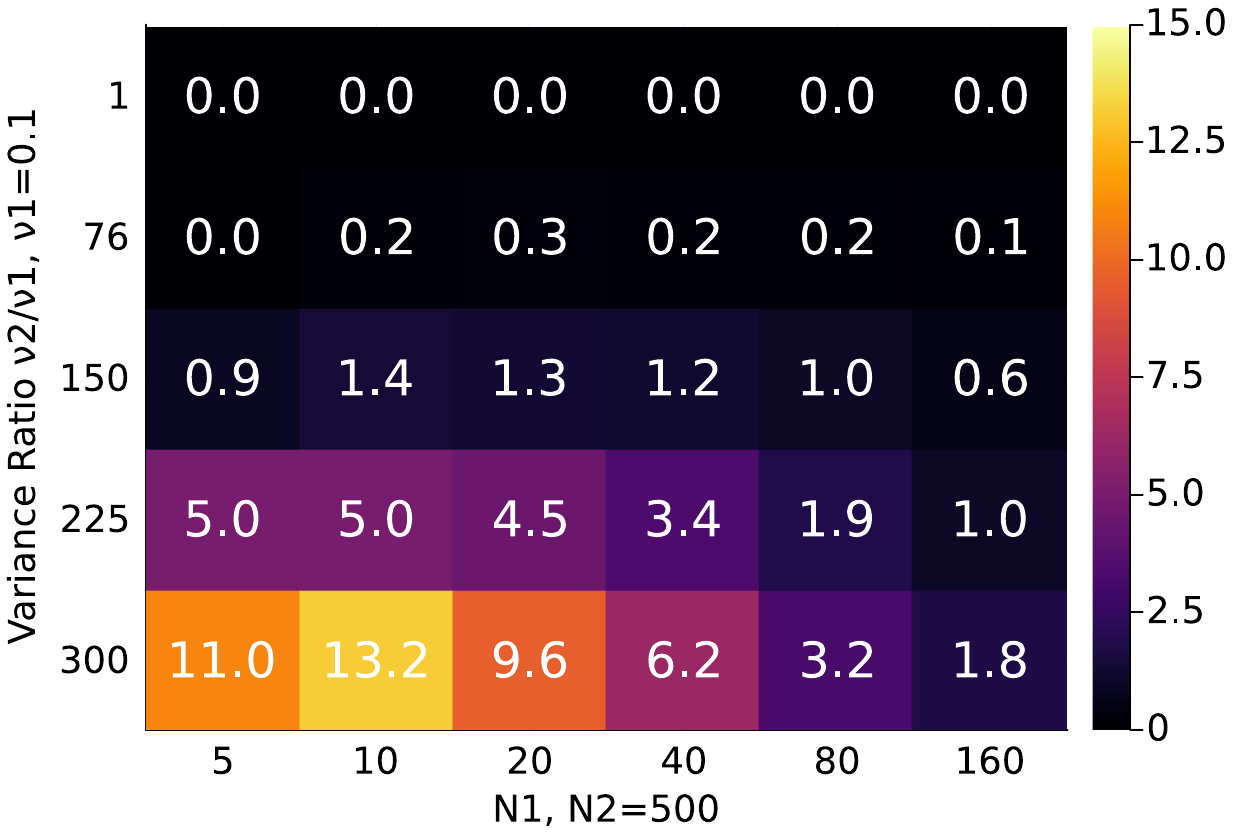}
    \caption{Percentage difference (\%) of ALPCAHUS clustering error
    subtracted from EKSS while good data amount varies.}
    \label{fig:clustering_good_data_experiment}
\end{figure}

\begin{figure}
    \centering
    \includegraphics[width=0.75\textwidth]{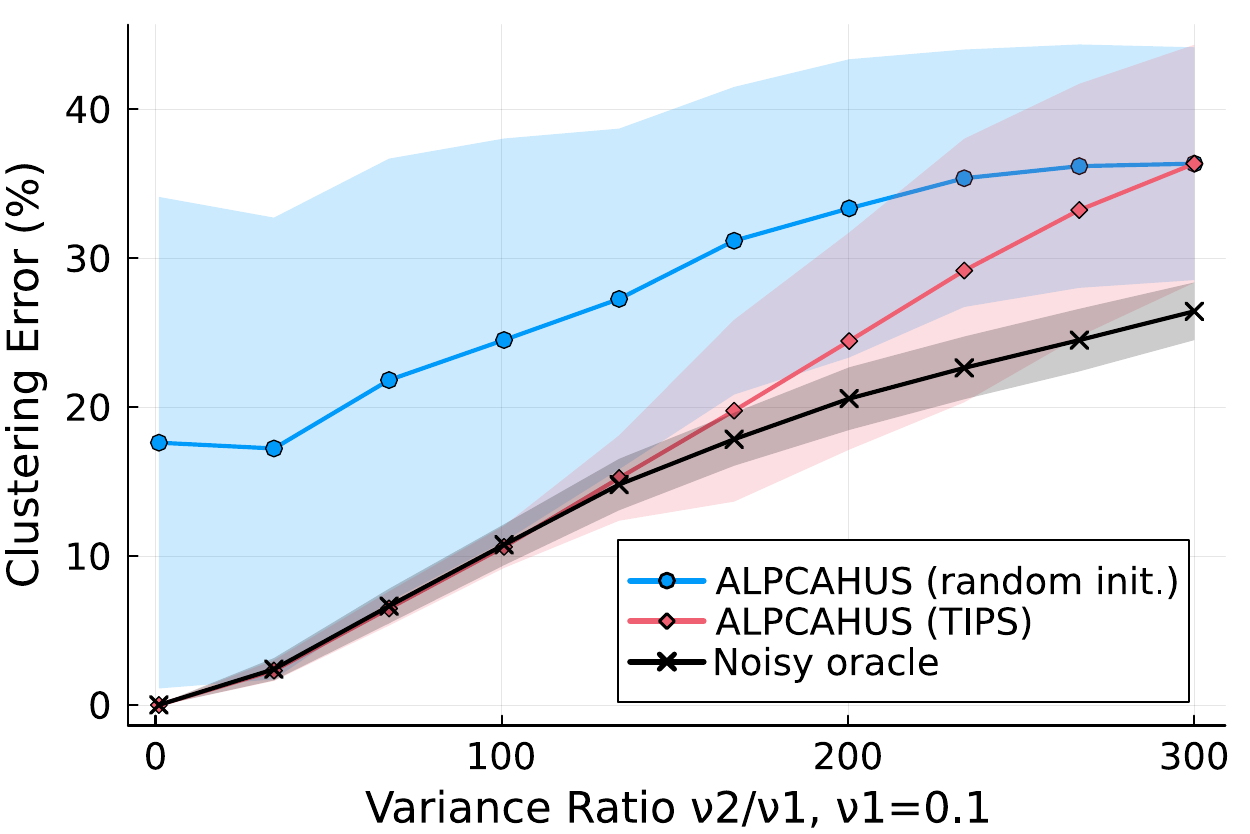}
    \caption{Clustering error (\%) for TIPS initialization scheme vs. random initialization
    for the ALPCAHUS method ($B=1$).}
    \label{fig:spectral_init}
\end{figure}

\begin{figure}[t]
    \centering
    \includegraphics[width=0.75\textwidth]{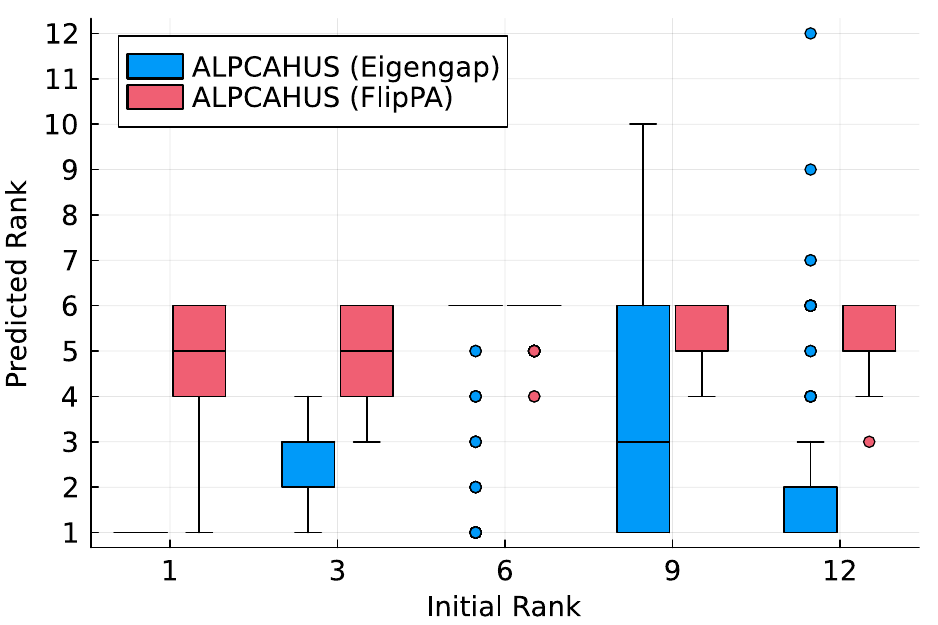}
    \caption{Adaptive rank estimation using eigengap heuristic and proposed FlipPA approach
    (true rank $d = 6$).}
    \label{fig:rank_estimation}
\end{figure}

\subsubsection{Clustering Error} 

\fref{fig:tab_clustering_landscape}
shows the clustering quality of these algorithms
for synthetic heteroscedastic data.
\fref{fig:clustering_heatmaps} complements \fref{fig:tab_clustering_landscape}
by exploring the complete heteroscedastic landscape
with a visual representation that is easier to understand.
We define a method called ``noisy oracle''
where an oracle  
uses the true cluster assignments to apply PCA only on the low noise data for each cluster.
Using these subspace basis estimates of each cluster,
the oracle performs cluster assignments using \eref{eq:cluster_update}. 
This oracle reference provides a kind of lower bound on realistic clustering performance
since it can approximate subspace bases separately given the true labeling. 

ALPCAHUS ($B=1$) with TIPS initialization
achieved lower clustering error than KSS with TIPS initialization. 
For the ensemble methods ($B>1$),
both EKSS and ALPCAHUS significantly improved.
However, EKSS still had higher clustering error in more heteroscedastic regions.
Meanwhile, ALPCAHUS remained very close to the noisy oracle,
indicating it more accurately estimated the underlying true labeling.
Compared to other methods like TSC and ADSSC,
ALPCAHUS generally outperformed them.
Overall, the ensemble version of ALPCAHUS was generally more robust against heteroscedasticity
compared to other subspace clustering algorithms.
To summarize, the effects of data quality and data quantity is explored 
on the heteroscedastic landscape. To our knowledge, this is the first systematic analysis of heteroscedasticity effects on subspace clustering quality in the literature.

\begin{figure*}
\centering
\subcaptionbox{Subspace clustering results for various methods. \label{fig:astro_cluster}}
{
\includegraphics[width=0.45\linewidth]{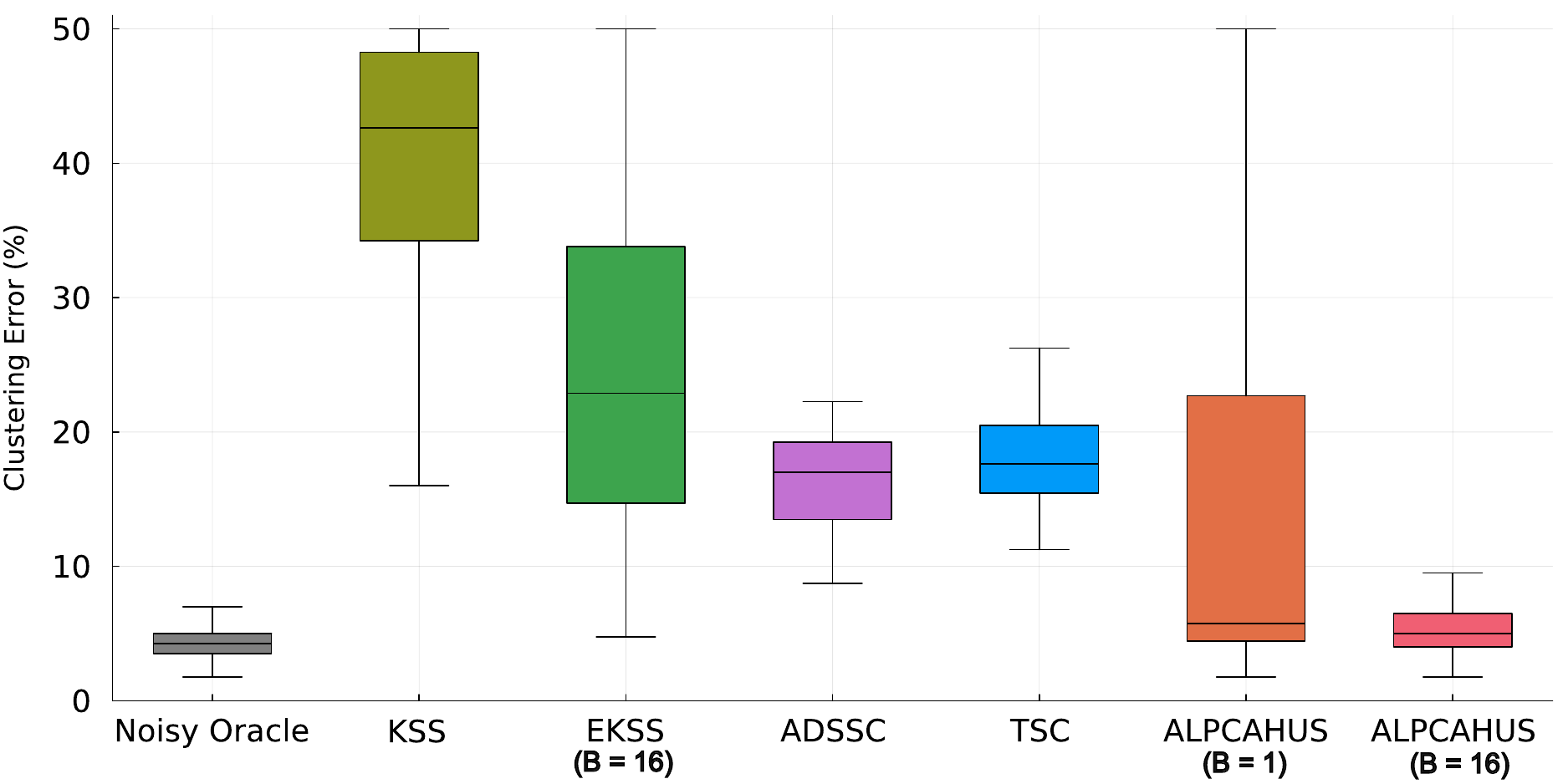}    
}
\hfil
\subcaptionbox{Average subspace affinity error results for various methods.
\label{fig:astro_cluster_subspaces}}
{
\includegraphics[width=0.45\linewidth]{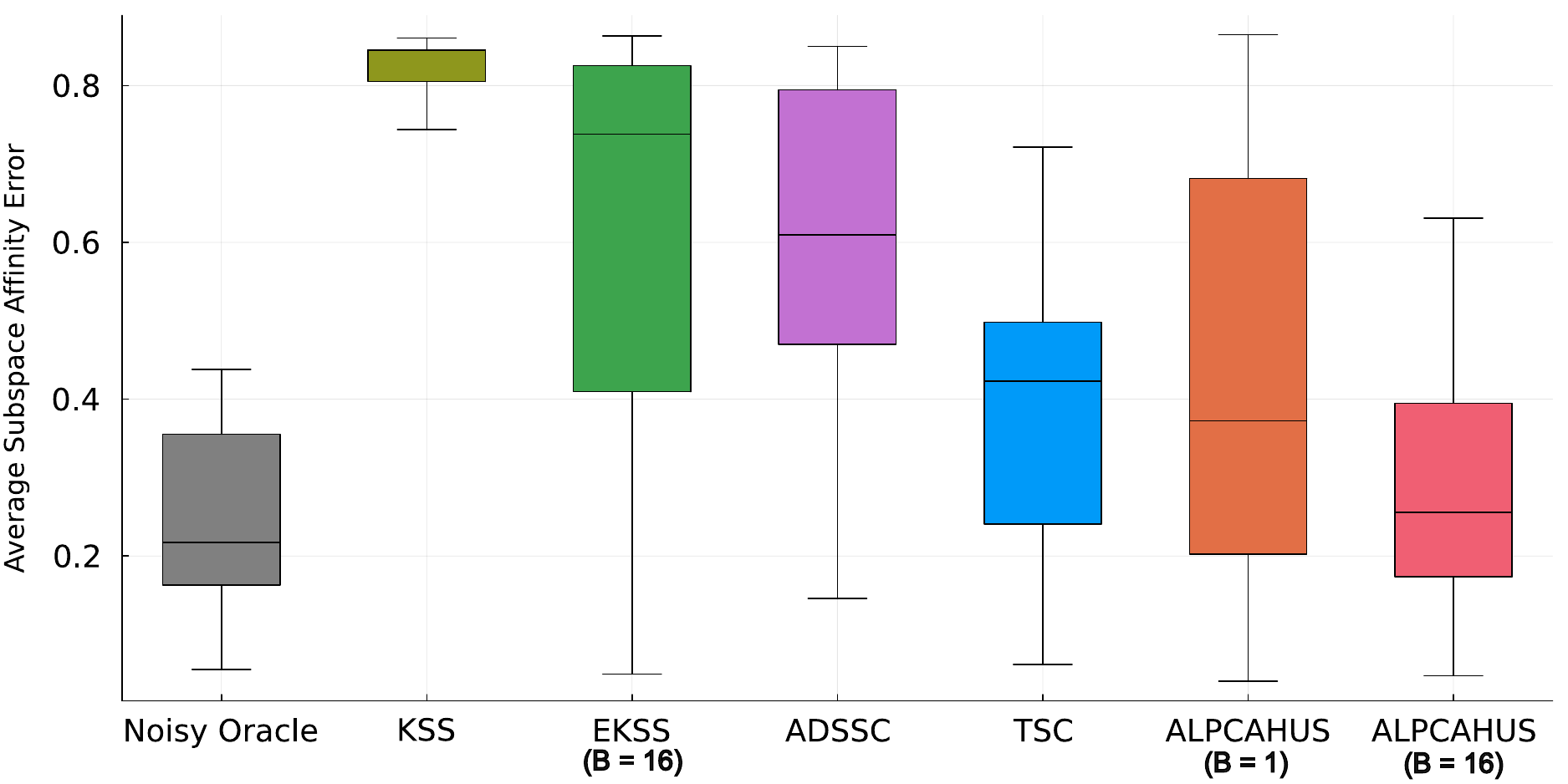}
}

\caption{Experimental results of quasar flux data for subspace clustering and learning. Methods involving a single run, i.e., KSS and ALPCAHUS ($B=1$), use the TIPS initialization scheme.}
\label{fig:quasar_clustering}
\end{figure*}

\subsubsection{Effects of Good Data} 

Additionally, the effects of good data quantity
on the subspace clustering problem is explored to understand at what point it becomes advantageous to use ALPCAHUS over the KSS method.
The following parameters are fixed $\{\nu_1 = 0.1, N_2 = 500\}$ and the following vary $\{N_1, \nu_2\}$. 
The reporting metric is the percentage difference of EKSS clustering error (\%) - ALPCAHUS clustering error (\%),
i.e., $\text{error}_{\text{EKSS}} - \text{error}_{\text{ALPCAHUS}}$,
meaning higher values indicate that ALPCAHUS is better than EKSS.  
\fref{fig:clustering_good_data_experiment}
shows that ALPCAHUS performed better than EKSS even up to $N_1 = 160$
which represents about 33\% of the total data.
ALPCAHUS never performed worse than EKSS
but the advantage gap narrowed as $N_1$ increased. 
Thus, there is a wide range of conditions
for which ALPCAHUS is preferable to homoscedastic methods.

\subsubsection{Clustering Initialization}

\sref{sec:method_init} proposed using the TIPS initialization scheme
in the heteroscedastic context
since the dot product metric used to construct the affinity matrix in \eref{eq:tips_metric} 
is provably robust to heteroscedastic noise. 
We fixed the following parameters $\{N_1,N_2\}$ and varied $\nu_2$ while keeping $\nu_1=0.1$.
The noisy oracle result is included to establish a realistic performance baseline.
\fref{fig:spectral_init}
shows that TIPS initialization ALPCAHUS $(B=1)$ outperformed random initialization
for the non-ensemble ALPCAHUS method $(B=1)$ across the entire heteroscedastic landscape.
Thus, TIPS should always be used for the non-ensemble version for higher clustering accuracy.
The ensemble version of ALPCAHUS $(B > 1)$ is not included in these results because the ensemble process inherently takes advantage of random initialization
to achieve a different labeling for each trial.

\subsubsection{Rank Estimation of Clusters} 

In \sref{sec:method_rank}, a random sign flipping method is proposed 
to adaptively find and shrink the subspace dimension of clusters when subspace dimension is unknown.
In \fref{fig:rank_estimation}, 
synthetic subspaces are generated with true rank $d = 6$
to explore how the initial rank parameter affects the ability
to estimate the true rank over 100 trials.
This approach is compared against the eigengap heuristic
by using ALPCAHUS with both adaptive rank schemes
and report the estimated rank values from the final clustering.
The eigengap approach consistently underestimated the rank
regardless of the initial value,
except when given the true rank value $d = 6$.
Our sign flipping approach provided much better rank estimates
with smaller variances between trials.

\subsection{Real Data Experiments}
\subsubsection{Quasar Flux Data}
\label{sec:experiments_quasar}
\begin{figure}
\centering
\includegraphics[width=0.8\textwidth]{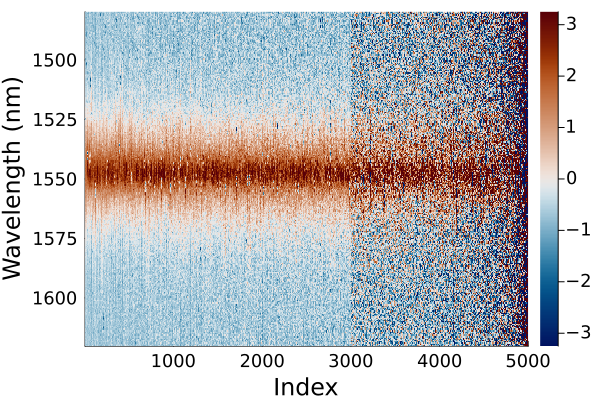}
\caption{Sample data matrix of quasar flux measurements across wavelengths
for each column-wise sample.}
\label{fig:astro_spectra}
\end{figure}

Quasar spectra data is downloaded from the Sloan Digital Sky Survey (SDSS) Data Release 16 \cite{sdss} using the DR16Q quasar catalog \cite{sdss2}.
Each quasar has a vector of flux measurements across wavelengths
that describes the intensity of observing that particular wavelength.
In this dataset, the noise is heteroscedastic
across the sample space (quasars) and feature space (wavelength),
but for these experiments, we focused on a subset of data that is homoscedastic across wavelengths and heteroscedastic across quasars.
The noise for each quasar is known
given the measurement devices used for data collection \cite{sdss}.
In \fref{fig:astro_spectra},
a subset of the spectra data is shown for illustrative purposes
to compare the visual differences in data quality.
The preprocessing pipeline for the data included
(filtering, interpolation, centering, and normalization)
based on the supplementary material of Ref. \cite{optimal_pca}.
Clusters are formed in these experiments by considering quasars with different properties, namely, two quasar groups ($K=2$) with different redshift values ($z_1 = 1.0-1.1, z_2 = 2.0-2.1$).
Additionally, for the second group,
only broad absorption lines (BAL) type quasar data is collected.
Since the downloaded data is queried separately,
clustering association is known, 
meaning one can compute clustering error for comparison purposes. 
A training set (5000 samples) is set aside to learn any model parameters
and the rest (5000 samples) is used for a test dataset. 
Trials are formed by randomly selecting 400 quasar spectra samples per group
and running clustering algorithms for 50 total trials.
For rank estimation, the noisy oracle approach is used, given the known cluster labels, to find that there is no improvement to clustering quality for rank values greater than $\hat{d}=3$,
so this value is used for any applicable algorithms.

\fref{fig:astro_cluster}
shows clustering error for this quasar spectra multi-cluster data.
The ensemble version of ALPCAHUS was very close in clustering quality
to the noisy oracle,
suggesting that it accurately learned the subspace bases while clustering the data groups.
The non-ensemble version of ALPCAHUS ($B=1$) 
had large variances in clustering error,
indicating some challenges in getting close to the optimal cost function minima with this data.
However, based on median values, it still managed to get a lower error than KSS, ADSSC, and TSC.
Additionally, \fref{fig:astro_cluster_subspaces}
shows the average subspace affinity error of these methods
after applying LR-ALPCAH to the clusterings for all methods.
ALPCAHUS better learned the subspace bases than the other methods when $B=16$, which was chosen as the smallest value that had small run-to-run variances while not taking significantly longer to compute. Both of these results show the benefit of developing heteroscedastic-based algorithms
in the subspace clustering context.
\tabref{fig:tab_clustering_astro}
reports median time complexity and memory requirements
along with mean clustering error and mean subspace error.
In this quasar example,
the ensemble method gets close in time to the non-ensemble methods
due to our multi-threaded implementation.
Yet, because of the multi-threaded implementation,
the memory requirements are larger for the ensemble method
as opposed to the non-ensemble method.
Overall, relative to other clustering methods,
ALPCAHUS is competitive time-wise at the cost of increased memory requirements.

\begin{figure*}
\centering
\subcaptionbox{Ground truth labels for \textit{Indian Pines} dataset. NC means no class available (background). \label{fig:pines_labels}}{%
\includegraphics[width=0.48\columnwidth]{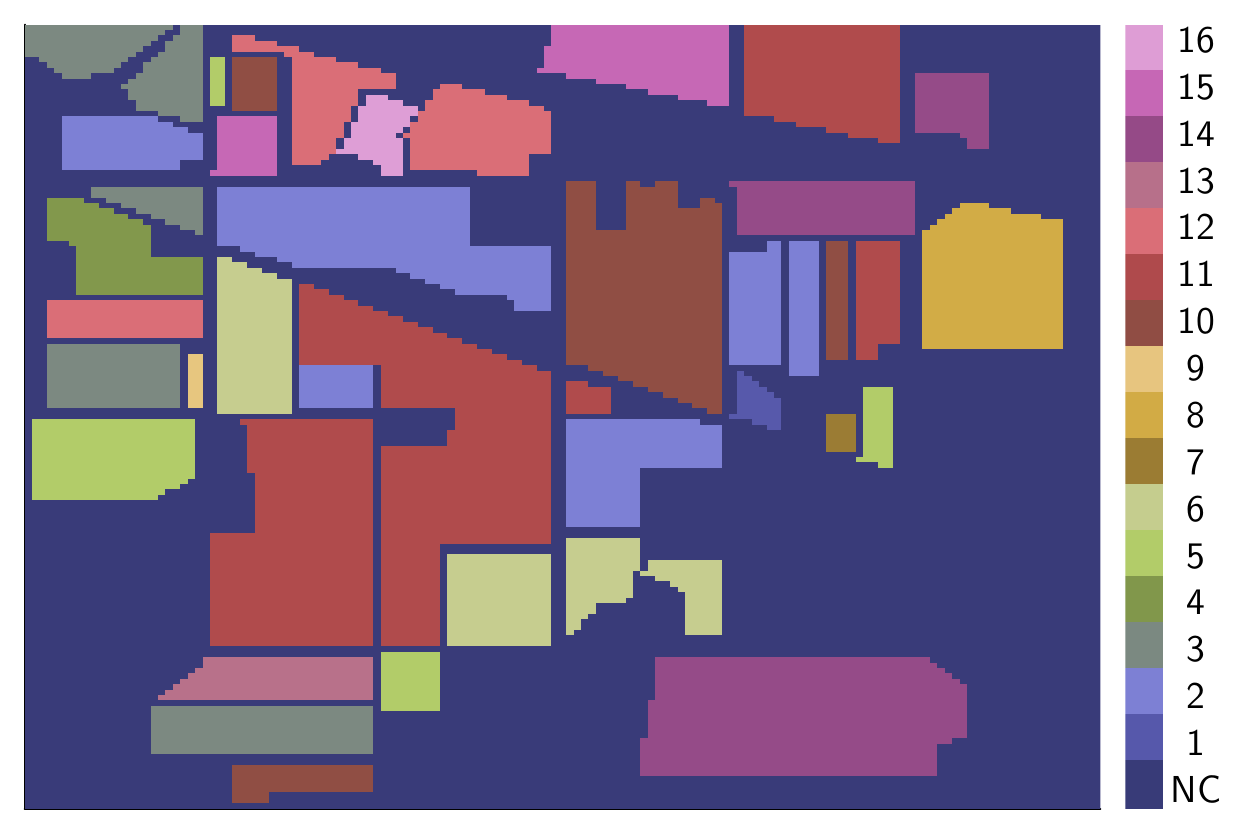}
}
\hfil
\subcaptionbox{$K$-means results with clustering error = 77\% and mIOU = 16\%. \label{fig:pines_kmeans}}{%
\includegraphics[width=0.48\columnwidth]{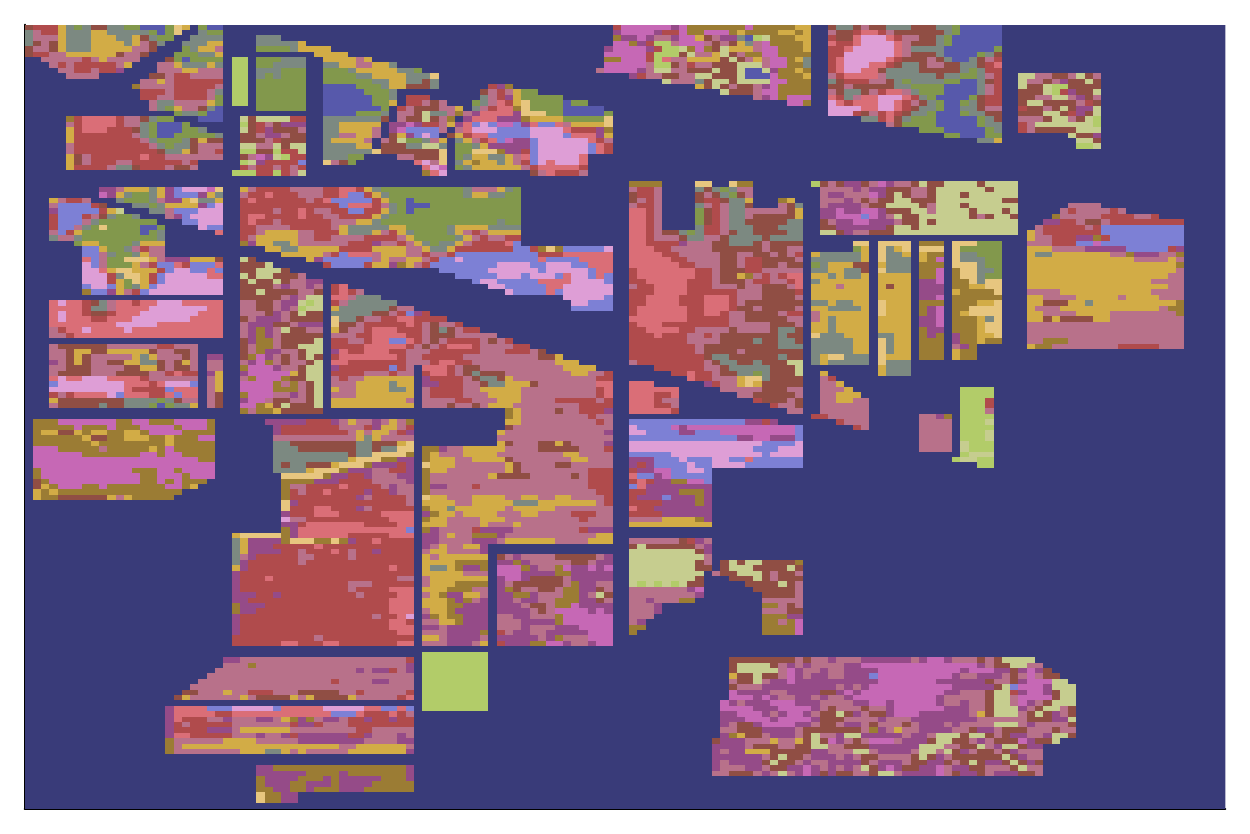}
}

\bigskip

\subcaptionbox{EKSS ($B=32, T=3, q=K$) results with clustering error = 64\% and mIOU = 27\%. \label{fig:pines_ekss}}{%
\includegraphics[width=0.48\columnwidth]{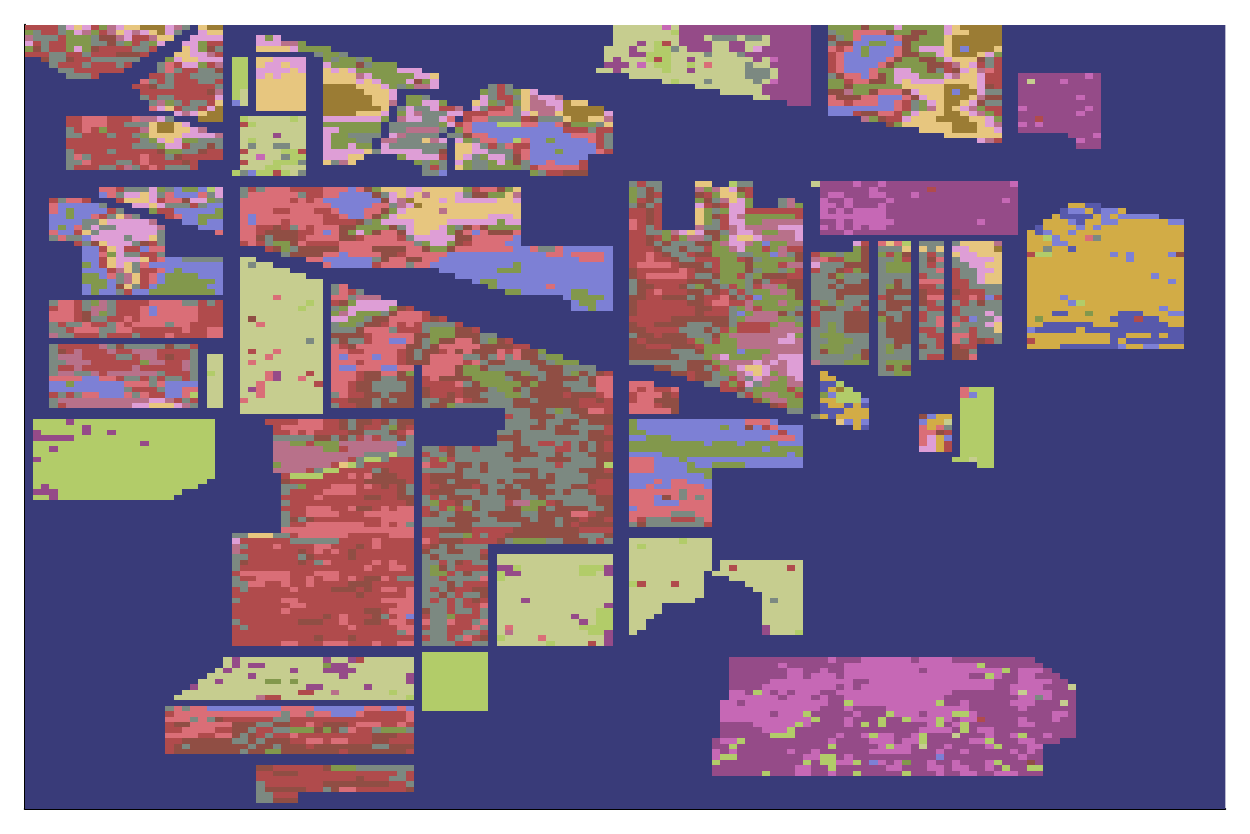}
}
\hfil
\subcaptionbox{ALPCAHUS ($B=32, T=3, q=K$) results with clustering error = 53\% and mIOU = 31\%. \label{fig:pines_alpcahus}}{%
\includegraphics[width=0.48\columnwidth]{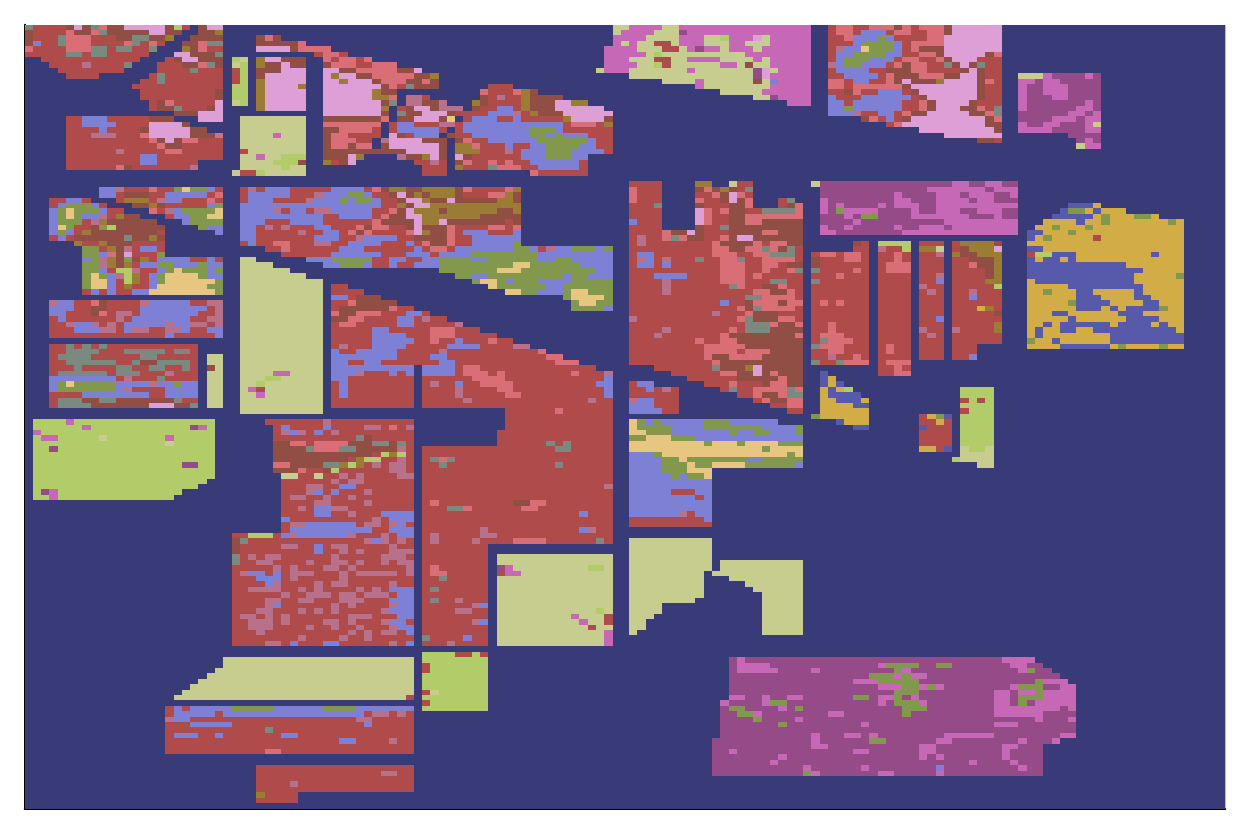}
}

\caption{Experimental results of \textit{Indian Pines} HSI data in a subspace clustering context. Results reported are clustering error and mean IOU (intersection over union) for each algorithm.}
\label{fig:pines}
\end{figure*}

\subsubsection{Indian Pines Data}
\label{sec:pines}

Hyperspectral image (HSI) segmentation data is explored in this section, specifically Indian Pines \cite{pines}. This image of size $145 \times 145$ contains $D=200$ reflectance bands for each pixel that is around the $0.4 - 2.5 \mu m$ wavelength. HSI data is known to be very noisy due to thermal effects, atmospheric effects, and camera electronics, leading to works that study per-pixel noise estimation in hyperspectral imaging \cite{hyperspectral1}. In total, there are $K = 16$ classes composed of alfalfa, corn, grass, wheat, soybean, and others. There is one additional class (background) that is withheld from clustering as commonly done in other works \cite{collas2021probabilistic}. Because of this, the data contains $N = 10,249$ samples instead of $N = 145 \times 145$ samples. In other works, researchers have found that $\hat{d} = 5$ is an appropriate rank parameter since it covers $95\%$ of the cumulative variance \cite{collas2021probabilistic}. A subset of the data matrix $Y \in \mathbb{R}^{200 \times 10249}$ is split into $2500$ samples to learn model parameters. From this, the learned subspaces are applied on $Y$ to cluster all data so that the heatmaps can be compared against the ground truth. The ground truth image is found in \fref{fig:pines_labels}. For perspective, randomly guessing would lead to 94\% clustering error due to $K=16$ clusters. For consistency, the Hungarian algorithm is again applied to the clustering results in \fref{fig:pines} to make it easier to compare against the ground truth label image in \fref{fig:pines_labels}. 

\begin{table}
\resizebox{0.97\columnwidth}{!}{%
\begin{tabular}{|l|r|r|r|r|}
\hline
\multicolumn{1}{|c|}{Methods}        & \multicolumn{1}{c|}{\begin{tabular}[c]{@{}c@{}}Time\\ (ms)\end{tabular}} & \multicolumn{1}{c|}{\begin{tabular}[c]{@{}c@{}}Memory\\ (MiB)\end{tabular}} & \multicolumn{1}{c|}{\begin{tabular}[c]{@{}c@{}}Mean\\ Subspace\\ Error \end{tabular}} & \multicolumn{1}{c|}{\begin{tabular}[c]{@{}c@{}}Mean\\ Clustering\\ Error (\%)\end{tabular}}\\ \hline
KSS & 149.3 & 53.0 & 0.80 & 38.5 \\ \hline
EKSS (B=16) & 181.4 & 212.6  & 0.62 & 23.8     \\ \hline
ADSSC & 125.7 & 17.5   & 0.62 & 21.1 \\ \hline
TSC & \textbf{32.3} & \textbf{16.5} & 0.40 & 18.2 \\ \hline
ALPCAHUS (B=1) & 148.5 & 126.7 & 0.43 & 14.4     \\ \hline
ALPCAHUS (B=16) & 210.7 & 678.2 & \textbf{0.28} & \textbf{6.3} \\ \hline
\end{tabular}
}
\caption{Subspace clustering results on quasar flux data. The KSS and ALPCAHUS ($B=1$) methods use TIPS initialization.}
\label{fig:tab_clustering_astro}
\end{table}

$K$-means is included in \fref{fig:pines_kmeans} to illustrate the difficulty of the problem. ALPCAHUS results are shown in \fref{fig:pines_alpcahus} next to EKSS results in \fref{fig:pines_ekss}. Clustering error is reported for these algorithms along with mean intersection over union (IOU) to measure the similarity between the images. ALPCAHUS achieved the lowest clustering error and highest mIOU values (53\%, 31\%) relative to the other approaches such as $K$-means (77\%, 16\%) and EKSS (64\%, 27\%). We note that our ALPCAHUS results are similar to Ref. \cite{collas2021probabilistic} even though, in that work, the authors treat the noise as homoscedastic and the signal itself as heteroscedastic. Further, the EKSS results on this dataset are similar to the homoscedastic PCA approach used in Ref. \cite{collas2021probabilistic}. This indicates some utility in modeling heteroscedasticity in hyperspectral images. Yet, the reflectance bands themselves also appear to be heteroscedastic with some bands being noisier than others \cite{hyperspectral2}. Thus, developing a method that is doubly heteroscedastic with respect to both the samples and features is an interesting direction for future work. 

For reference, the state of the art result is about 10\% misclassification rate in a \textit{classification} setting (i.e., not clustering) \cite{indianpines_class_sota}. In a clustering setting, recent works such as Ref. \cite{indianpines_clustering_sota} have achieved a 40\% clustering error which is 13\% lower than our results. We do not claim state of the art results with this HSI data, only that heteroscedastic union of subspace modeling improved results over homoscedastic union of subspace modeling in finding reflectance band groups with similar characteristics.

\newpage
\section{Conclusion}
\label{paper:conclusion}
This paper proposed ALPCAHUS,
a subspace clustering algorithm
that can find subspace clusters whose samples contain heteroscedastic noise.
For future work, a union of manifolds generalization by deep learning could be useful.
For example, Alzheimer's disease patients often have resting state functional MRI activations different than cognitively normal individuals \cite{function_mri_disease} \cite{functional_mri_vae},
so one could consider these different classes
to belong to different manifolds in the ambient space.
Another direction of future work could be to consider heteroscedasticity across the feature space instead of the sample space. For example, one might be interested in biological sequencing of different species with similar genes
where the gene marker counts naturally follow heteroscedastic distributions \cite{bio_hetero}. 
Another interesting topic would be to explore \tref{thm:alpcahus} in greater detail.
This theorem simply states that ALPCAHUS generates a sequence of cost function values that converge; it says nothing about recovery guarantees or the quality of that solution.
Thus, provably showing how the basin of attraction changes as heteroscedastic data is introduced would lead to a greater understanding of heteroscedasticity in a subspace clustering context. Additionally, per \sref{appendix:balancedcluster}, extending the method to handle extremely imbalanced cluster sizes would be another interesting avenue. Further, a theoretical framework for rank estimation for real applications would be complimentary to the empirical approach presented here. These generalizations are nontrivial so they are left for future work.

While our implementation of ALPCAHUS benefited from parallelization,
it needed more computation time and memory than TSC.
Because of this, there is an opportunity to further optimize our math modeling
to remove certain matrix multiplication operations
and instead use single vector dot products to reduce memory.
Further, since our subspace basis step requires an SVD computation
for an initial low-rank estimate, perhaps the Krylov-based Lanczos algorithm \cite{fast_svd} can be used to further reduce time and memory for each trial leading
to more significant improvements in memory usage and time. These additions can be fruitful in big data settings. 
Overall, ALPCAHUS was more robust towards heteroscedastic noise than other clustering methods, making it easier to identify correct clusterings under this kind of heterogeneity model.  

\newpage
\section{Acknowledgments}
This work was supported in part by the National Science Foundation (NSF) under NSF CAREER Grant CCF-1845076, NSF Grant CCF-2331590, and by the U.S. National Institutes of Health (NIH) under Grant R21 AG082204. 
We thank David Hong for suggesting heteroscedastic data applications
such as astronomy spectra that are used in this work.
\section{Appendix}
\subsection{ALPCAHUS Convergence Experiment}
\label{appendix:convergence}
\begin{figure}
    \centering
    \includegraphics[width=0.97\columnwidth]{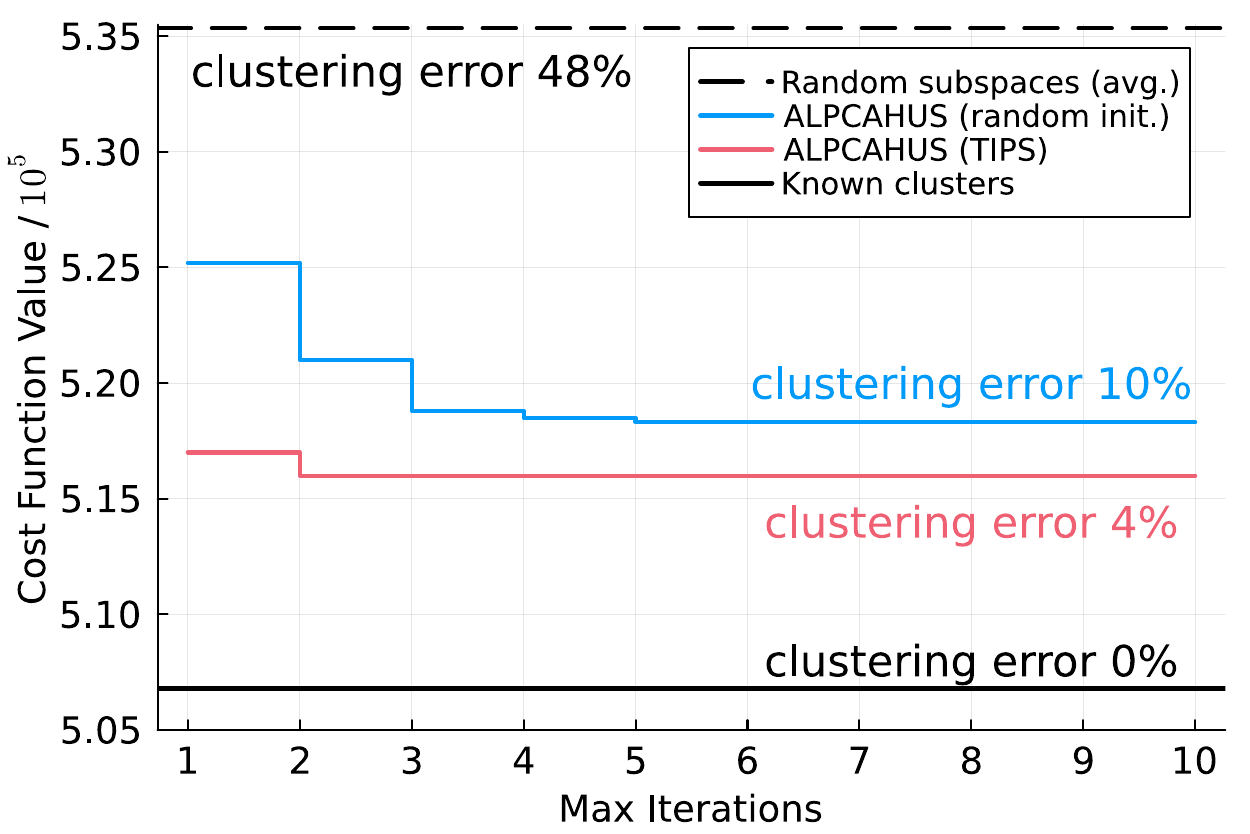}
    \caption{ALPCAHUS ($B=1$) cost function value convergence plot to corroborate the theorem in \tref{thm:alpcahus}.}
    \label{fig:convergence}
\end{figure}

This section focuses on providing empirical verification of the ALPCAHUS convergence theorem in \tref{thm:alpcahus}. The quasar spectra flux data from SDSS (DR16Q catalog) is used in this experiment. The noise variance threshold parameter $\alpha$ is set to $10^{-6}$ to lower bound the noise variance estimation. Further, as shown in \tref{thm:alpcahus}, it is necessary to use the cluster reassignment criteria in \eref{eq:cluster_update} that rejects repeated assignments of points to prevent cycling. For simplicity, only one base clustering ($B=1$) is used in this experiment to compare a randomly initialized ALPCAHUS trial against the TIPS initialized variant. This variant is only possible when $B=1$ as discussed in the paper. In \fref{fig:convergence}, the cost function value is plotted over iterations for both versions of ALPCAHUS.

Both upper and lower bound cost function reference values are provided in \fref{fig:convergence}. The upper bound is generated by using randomly initialized subspaces, and by applying \eref{eq:cluster_update} to get cluster labels. From this, the cost function value is computed by \eref{eq:alpcahus}. Likewise, a lower bound is generated by using the known cluster labels, and by estimating the subspaces and noise variances by \eref{eq:lr-alpcah}. As observed in \fref{fig:convergence}, ALPCAHUS converges relatively quickly with only a few iterations.
Note that in this figure, the stopping criteria in line \eqref{alg:finite_iterations} of \aref{alg:alpcahus} is purposely ignored. In other words, ALPCAHUS is run for a fixed number of iterations. However, as clearly demonstrated, convergence is achieved earlier for both versions of ALPCAHUS before reaching the maximum number of iterations. This indicates the usefulness of the stopping criteria to speed up the algorithm in various applications.

\subsection{Balanced Cluster Assumption}
\label{appendix:balancedcluster}
The ensemble extension of ALPCAHUS relies on forming an affinity matrix in \eref{eq:coassociation_matrix} that is thresholded by \eref{eq:threshold_rows} and \eref{eq:threshold_columns}. Then, spectral clustering is applied on said affinity matrix which depends on \eref{eq:ratio_cut}. This spectral clustering operation, by the normalized cut function in \eref{eq:ratio_cut}, assumes balanced cluster partitions to avoid trivial solutions such as finding a small cut that leads to a single point belonging to its own cluster and everything else being in another cluster. Thus, the proposed ALPCAHUS method may not be ideal in situations where imbalanced clusters exist. 

While one could use other clustering methods for imbalanced data such as Ref. \cite{pet_turtle} or Ref. \cite{subspace_clustering_imbalanced}, these methods are not ideal in heterogeneous settings as they do not account for heteroscedasticity in the data. Instead, we recommend using the partition cut (``Pcut'') method introduced in Ref. \cite{spectral_clustering_imbalanced} that performs spectral clustering-like operations on a rank-modulated degree graph as a replacement for spectral clustering which uses the normalized cut function (``Ncut'') in \eref{eq:ratio_cut}. This would allow for ALPCAHUS to be more robust towards imbalanced cluster size situations. Furthermore, the non-ensemble version of ALPCAHUS in \eref{eq:alpcahus}, may also suffer in the imbalanced cluster regime if the TIPS-based initialization scheme in \sref{sec:method_init} is used. This is due to the spectral clustering operation in TIPS to initialize cluster labels for ALPCAHUS. Therefore, even when $B=1$, it is also recommended to apply the ``Pcut'' method in Ref. \cite{spectral_clustering_imbalanced} on the affinity matrix formed by \eref{eq:aij-tips}.

\subsection{ALPCAHUS Parameters}
\label{appendix:parameters}
For \aref{alg:alpcahus}, the input data matrix $Y \in \mathbb{R}^{D \times N}$ and the number of subspaces/clusters $K \in \mathbb{Z}^+$ are necessary inputs that are clearly data dependent. For the candidate subspace dimension $\hat{d}_k$, it may be the same across all clusters or vary, and it is either known via domain knowledge or can be reasonably estimated by the procedure outlined in \sref{sec:method_rank}. If it must be estimated, it is recommended to start over-parameterized rather than under-parameterized. A scree plot of $Y$ can give an indication of sufficient rank for initialization to ensure an over-parameterized state. Further, $\hat{d}_k$ may also be treated simply as a hyperparameter for cross validation purposes.

For the threshold parameter $q \in \mathbb{Z}^+$, this can be set to the subspace dimension as shown in Ref. \cite{l2graph_subspace_clustering} and explained in \sref{sec:method_ensemble}, or simply treated as a hyperparameter to be optimized during cross validation; this value is dataset dependent. The number of base clusterings $B \in \mathbb{Z}^+$ is either fixed to $B=1$ for the non-ensemble version of ALPCAHUS, or set as high as compute time and memory allows for the ensemble version. In other words, more trials make it easier to identify the underlying relationships between similar data samples. If $B=1$, then it is recommended to use the TIPS-based initialization scheme from \sref{sec:method_init}.

Since each subspace basis update involves LR-ALPCAH \cite{alpcah_journal}, the $T_1 \in \mathbb{Z}^+$ parameter specifies the number of LR-ALPCAH iterations. For $T_1$, it is recommended to keep some low value, e.g., we set $T_1=5$ since there is no point in converging perfectly as the cluster labels will get updated in the next reassignment anyway at the $t_2+1$ iteration. For the non-ensemble version, it is recommended to do as many alternating updates $T_2 \in \mathbb{Z}^+$ as necessary to trigger the stopping criteria in line \eref{alg:finite_iterations} of \aref{alg:alpcahus}. However, for the ensemble version of ALPCAHUS, we find that a low value such as $T_2=3$ works very well in practice due to the larger number of trials $B \gg 1$. This value agrees with \fref{fig:convergence} as ALPCAHUS converges in just a few iterations anyway. For more detailed and practical code-oriented function usage, refer to the documentation at \url{github.com/javiersc1/ALPCAHUS}.

\bibliographystyle{IEEEtran}
\bibliography{setup/ref.bib}

\newpage
\begin{IEEEbiography}[{\includegraphics[width=1in,height=1.25in,clip,keepaspectratio]{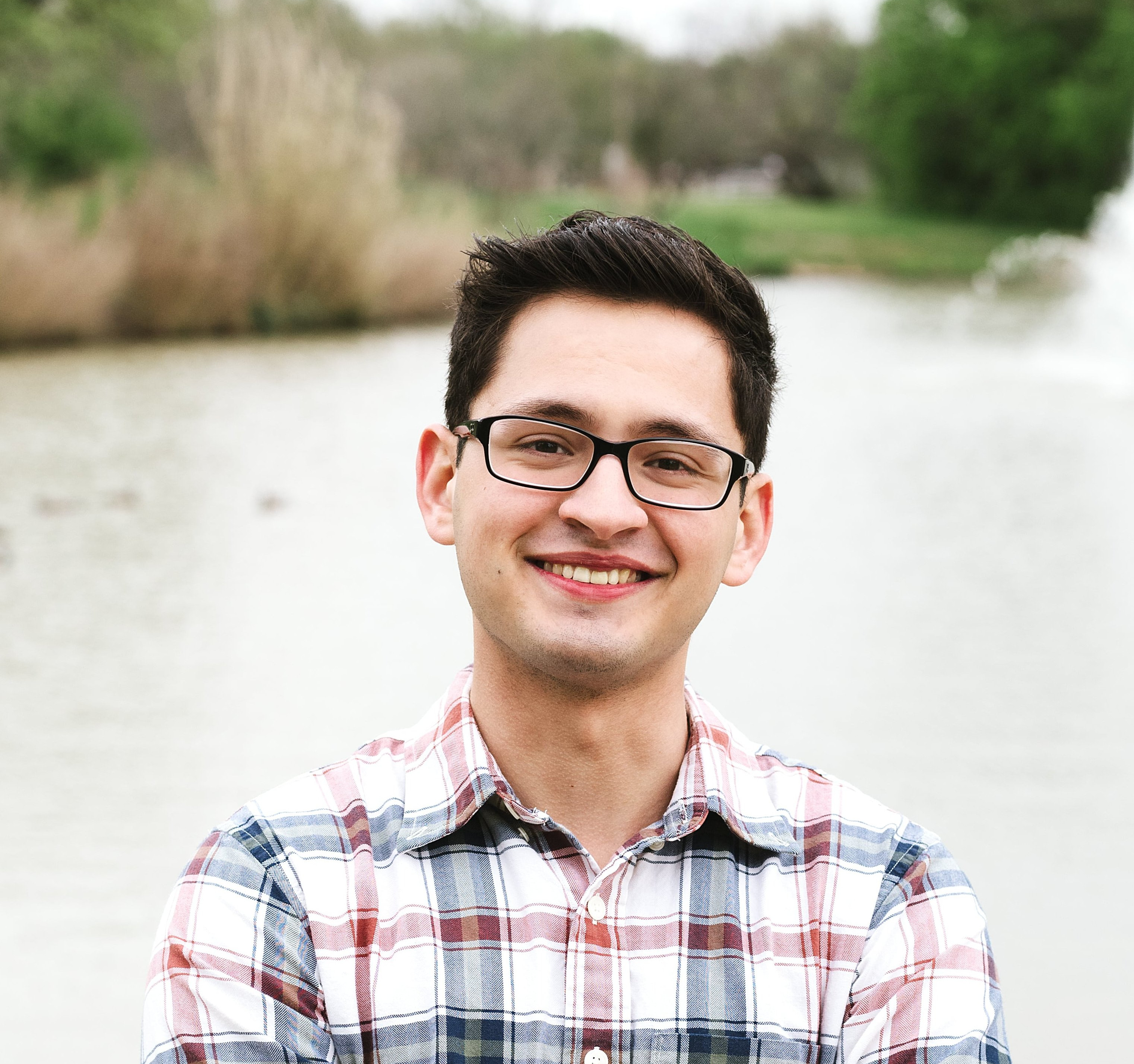}}]{Javier Salazar Cavazos}
(Graduate Student Member, IEEE) received dual
B.S. degrees in electrical engineering and mathematics 
from the University of Texas at Arlington, Arlington, TX, USA, in 2020, and the M.S. degree in electrical and computer engineering from the University of Michigan, Ann Arbor, MI, USA, in 2023. He is
currently working towards the Ph.D. degree in electrical 
and computer engineering working with Laura Balzano, Jeffrey Fessler, and Scott Peltier, all with the University of Michigan, Ann Arbor, MI, USA. 
His main research interests include signal and image processing, machine learning, deep learning, and optimization. Current work focuses on topics such as low-rank modeling, clustering, noisy data, and Alzheimer's disease diagnosis and prediction from functional MRI data.
\end{IEEEbiography}

\begin{IEEEbiography}[{\includegraphics[width=1in,height=1.25in,clip,keepaspectratio]{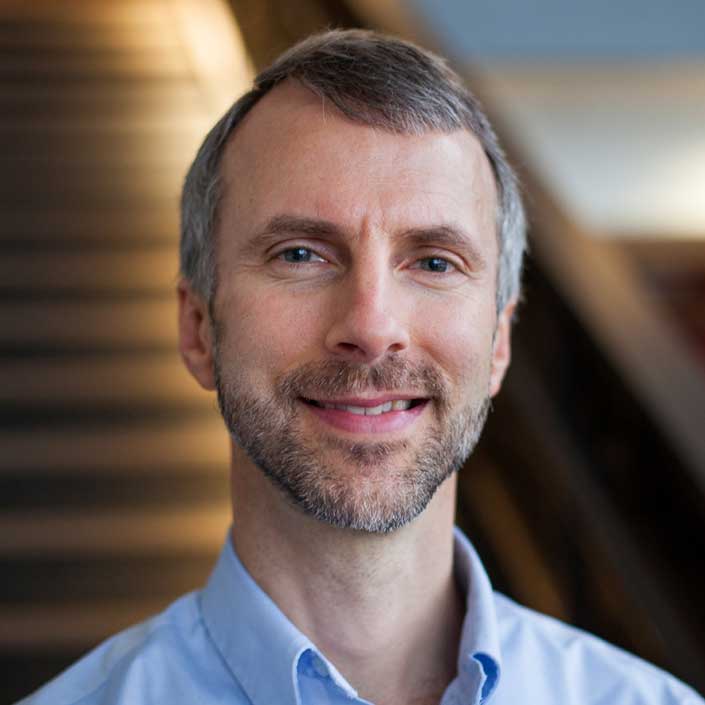}}]{Jeffrey A. Fessler}
(Fellow, IEEE) received the B.S.E.E. degree from Purdue University, West Lafayette, IN, USA, in 1985, the M.S.E.E. degree from Stanford University, Stanford, CA, USA, in 1986, and the M.S. degree in statistics and the Ph.D. degree in electrical engineering from Stanford University, in 1989 and 1990, respectively. From 1985 to 1988, he was a National Science Foundation Graduate Fellow with Stanford. Currently, he is the William L. Root Distinguished University Professor of EECS with the University of Michigan, Ann Arbor, MI, USA. His research focuses on various aspects of imaging problems, and he has supervised doctoral research in PET, SPECT, X-ray CT, MRI, and optical imaging problems. In 2006, he became a fellow of the IEEE for contributions to the theory and practice of image reconstruction.
\end{IEEEbiography}

\begin{IEEEbiography}[{\includegraphics[width=1in,height=1.25in,clip,keepaspectratio]{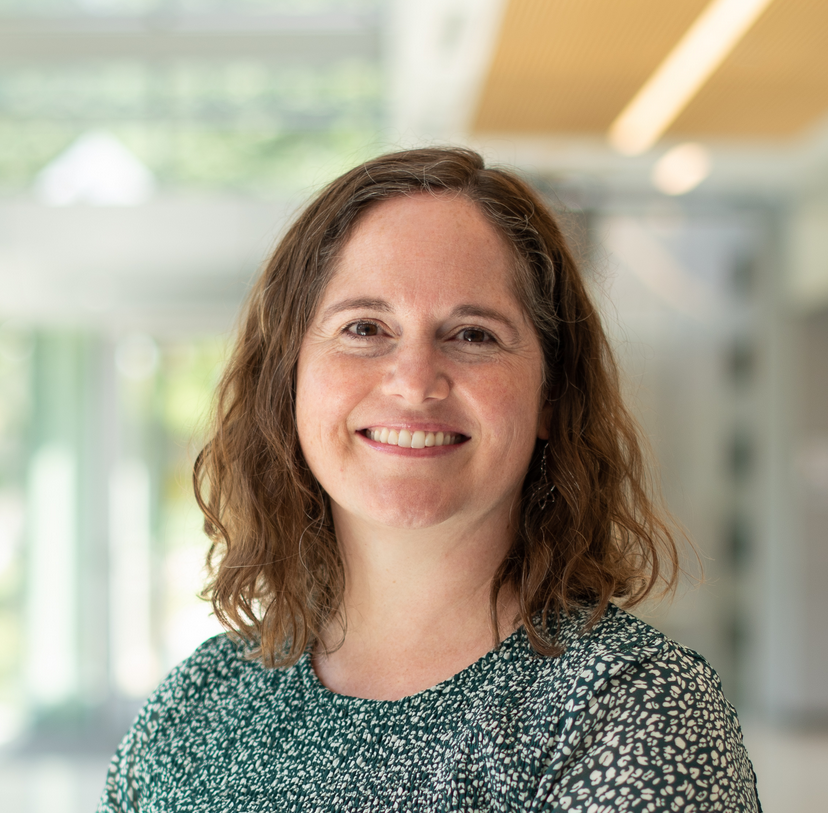}}]{Laura Balzano}
(Senior Member, IEEE) received the Ph.D. degree in ECE from the University of Wisconsin, Madison, WI, USA. Currently, she is an Associate Professor of electrical engineering and computer science with the University of Michigan, Ann Arbor, MI, USA. Her main research focuses on modeling and optimization with big, messy data – highly incomplete or corrupted data, uncalibrated data, and heterogeneous data – and its applications in a wide range of scientific problems. She is an Associate Editor for IEEE Open Journal of Signal Processing and SIAM Journal of the Mathematics of Data Science. She was a recipient of the NSF Career Award, the ARO Young Investigator Award, and the AFOSR Young Investigator Award. She was also a recipient of the Sarah Goddard Power Award at the University of Michigan.
\end{IEEEbiography}

\end{document}